\DocumentMetadata{}
\documentclass[sigconf]{acmart}
\usepackage[normalem]{ulem}
\useunder{\uline}{\ul}{}
\usepackage{algorithm}
\usepackage{algpseudocode}
\usepackage{amsmath}
\usepackage{amsfonts}
\usepackage{diagbox}
\usepackage{multirow}
\usepackage{xspace}
\usepackage{multirow}
\usepackage{bm}
\usepackage{amsmath, nccmath}
\usepackage{geometry}
\usepackage{enumitem}
\usepackage{subfigure}
\usepackage[skip=0pt]{caption}
\usepackage{multirow}
\usepackage{xcolor}
\usepackage{color, colortbl}
\usepackage{hyperxmp}
\usepackage{hyperref}
\definecolor{Gray}{gray}{0.95}

\AtBeginDocument{%
  \providecommand\BibTeX{{%
    \normalfont B\kern-0.5em{\scshape i\kern-0.25em b}\kern-0.8em\TeX}}}

\setcopyright{acmcopyright}
\copyrightyear{2023}
\acmYear{2023}
\acmDOI{XXXXXXX.XXXXXXX}

\acmConference[WSDM '24]{Make sure to enter the correct
  conference title from your rights confirmation email}{May 4th--8th,
  2024}{Mérida, Yucatán, Mexico}

\acmPrice{15.00}
\acmISBN{978-1-4503-XXXX-X/18/06}

\newcommand{\proj}{GAD-NR\xspace}

\begin{document}

\title [GAD-NR: Graph Anomaly Detection via  Neighborhood Reconstruction]{GAD-NR: Graph Anomaly Detection via \\ Neighborhood Reconstruction}

\author{Amit Roy}
\affiliation{
  \institution{Purdue University}}
\email{roy206@purdue.edu}

\author{Juan Shu}
\affiliation{
  \institution{Purdue University}}
\email{shu30@purdue.edu}

\author{Jia Li}
\affiliation{
  \institution{HKUST}}
\email{jialee@ust.hk}

\author{Carl Yang}
\affiliation{
  \institution{Emory University}}
\email{j.carlyang@emory.edu}

\author{Olivier Elshocht}
\affiliation{
  \institution{Sony}}
\email{Olivier.Elshocht@sony.com}

\author{Jeroen Smeets}
\affiliation{
  \institution{Sony}}
\email{jeroen.smeets@sony.com}

\author{Pan Li}
\affiliation{
  \institution{Georgia Institute of Technology, Purdue University}}
\email{panli@gatech.edu,   panli@purdue.edu}

\begin{abstract}
Graph Anomaly Detection (GAD) is a technique used to identify abnormal nodes within graphs, finding applications in network security, fraud detection, social media spam detection, and various other domains. A common method for GAD is Graph Auto-Encoders (GAEs), which encode graph data into node representations and identify anomalies by assessing the reconstruction quality of the graphs based on these representations. However, existing GAE models are primarily optimized for direct link reconstruction, resulting in nodes connected in the graph being clustered in the latent space. As a result, they excel at detecting cluster-type structural anomalies but struggle with more complex structural anomalies that do not conform to clusters. To address this limitation, we propose a novel solution called \proj, a new variant of GAE that incorporates neighborhood reconstruction for graph anomaly detection. \proj aims to reconstruct the entire neighborhood of a node, encompassing the local structure, self-attributes, and neighbor attributes, based on the corresponding node representation. By comparing the neighborhood reconstruction loss between anomalous nodes and normal nodes, \proj can effectively detect any anomalies. 
Extensive experimentation conducted on six real-world datasets validates the effectiveness of \proj, showcasing significant improvements (by up to 30\%$\uparrow$ in AUC) over state-of-the-art competitors.
The \href{https://github.com/anonymoususer437/GAD-NR}{\textcolor{cyan}{source code}} for \proj is openly available. Importantly, the comparative analysis reveals that the existing methods perform well only in detecting one or two types of anomalies out of the three types studied. In contrast, GAD-NR excels at detecting all three types of anomalies across the datasets, demonstrating its comprehensive anomaly detection capabilities.

\end{abstract}


\maketitle

\section{Introduction}

Anomaly Detection aims to identify entities that deviate significantly from the norm, which has been used for a variety of applications, such as revealing fraudulent or spam activity in social networks~\cite{wang2020atne, van2017gotcha, savage2014anomaly,yu2016survey,ye2015discovering, heard2010bayesian} and financial transactions networks~\cite{starnini2021smurf, dou2020enhancing, tang2022rethinking, elliott2019anomaly,wang2019semi, AntiBenford,pei2020subgraph}.

Unlike anomaly detection methods for tabular and time-series data, Graph Anomaly Detection (GAD)~\cite{akoglu2015graph, GraphAnomalySurvey,tang2023gadbench} poses additional challenges. Graph data is often multi-modal, containing information from both node/edge attributes and topological structures. This complexity makes it difficult to find a unified definition of anomalies for graph-structured data and to design a principled algorithm for detecting them.

\begin{figure}[!t]
\centering
\includegraphics[width=3.2in]{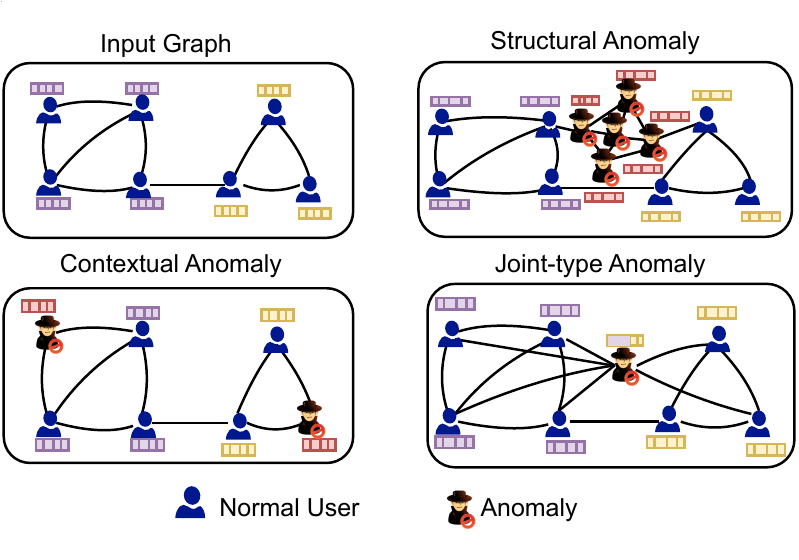}
\caption{\small Contextual anomalies are feature-wise different, structural anomalies form dense subgraphs in the network and joint-type anomalies connect with many nodes with different features. }
\vspace{-2em}
\label{fig:outlier_types}
\end{figure}

Due to the inherent multi-modality of graph-structured data, anomalies on graphs can be grouped into three categories: contextual, structural, and joint-type, as illustrated in Fig.~\ref{fig:outlier_types}. Contextual anomalies refer to nodes whose attributes are vastly different from those of regular nodes, such as spammers or fake account holders in social media networks~\cite{hu2014online,jia2017random,xiao2015detecting}. Structural anomalies refer to nodes with different connectivity patterns compared to other nodes, such as a group of malicious sellers exchanging fake reviews with super dense connections~\cite{wang2021bipartite} or bots retweeting the same tweet forming a densely connected co-retweet  network~\cite{hooi2016fraudar,feng2021botrgcn}. Joint-type anomalies are those that can only be identified by considering both attributes and connectivity patterns, such as a node that is sending a large number of phishing emails to users across different communities in an email network~\cite{pan2022semantic, karim2019comprehensive}. To identify all these types of anomalies, we need a powerful model to capture attribute information, connectivity patterns, and most importantly the correlation between them.

However, current GAD approaches~\cite{liu2022pygod, akoglu2015graph, GraphAnomalySurvey} only perform well to detect one or two types of these anomalies but not all of them. Some GAD approaches only leverage network structure, which cannot detect contextual anomalies. Examples include the methods to check centrality measures or clustering coefficients~\cite{maulana2020centrality,shashikala2015outlier},  based on factorization of the adjacency matrix~\cite{tong2011non}, and 
performing network clustering~\cite{SCAN}. Some approaches check the distribution of node features to detect anomalies~\cite{LOF, IF}, such as using the $k$-nearest neighbor algorithm on node features, to detect nodes that are isolated from others. These approaches fail to detect anomalies other than contextual anomalies.

Recently, autoencoders have been widely employed for anomaly detection~\cite{MLPAE,GCNAE,Dominant,DONE,AnomalyDAE}. The rationale is that autoencoders leverage neural networks to reduce the dimension of the data. Anomalies are often sparse in the data and hence such a data compression process tends to record only the principal part of the data and automatically exclude sparse anomalies.  Therefore, one can use the obtained compressed data representations to approximately reconstruct the normal data but not the anomalies. Monitoring the reconstruction loss can thus identify those anomalies from the normal data. For GAD,  Graph Auto-Encoders (GAEs) have been proposed to leverage Graph Neural Networks (GNNs)~\cite{GCN, GRAPHSAGE, GIN} to encode both graph structure and node attributes, which have recently been used to detect anomalies on graphs~\cite{GCNAE, Dominant, AnomalyDAE}. 

However, current GAE-based methods~\cite{Dominant, GCNAE, AnomalyDAE} often adopt a strategy of reconstructing direct links between nodes based on their representations, which brings the nodes close to each other in the latent space that are originally connected in the graph structure. Such a proximity-driven loss to reconstruct graph structures may be effective to detect structural anomalies that are inherently clustered together in the graph. 
However, they fail to detect joint-type anomalies that are not naturally clustered. Intuitively, joint-type anomalies rely on the entire neighborhoods for correct detection, because the information of which nodes are connected and the attributes on these neighboring nodes is useful for the detection.  

In this paper, we address the current limitation and propose a novel framework \textbf{G}raph \textbf{A}nomaly \textbf{D}etection via \textbf{N}eighborhood \textbf{R}econstruction (\textbf{\proj}). 
\proj extends a recently-proposed neighborhood reconstruction-based GAE model, namely NWR-GAE~\cite{NWRGAE} to address fundamental problems in GAD. Specifically, rather than using a proximity-driven loss to recover direct links, \proj imposes the dimension-reduced node representations to reconstruct the entire neighborhoods, i.e., the receptive fields that are encoded/compressed by GNNs into the node representations. Specifically, \proj aims to reconstruct the information of one's own attributes, its connectivity pattern, and the attributes of its neighboring nodes. By checking different types of reconstruction losses, \proj can detect all three types of anomalies.

The key novelty of \proj is that it is the first work that identifies neighborhood reconstruction as a powerful metric for GAD, which fundamentally differs from previous GAE models that adopt the metric of link reconstruction/prediction. 
Moreover, \proj also advances technical aspects of the backbone model NWR-GAE~\cite{NWRGAE} directly for GAD tasks, which yields substantial improvements in stability, scalability, and accuracy. Specifically, \proj adopts Gaussian approximation of neighbors' features distributions, which not only substantially reduces the computation cost of NWR-GAE but also avoids learning a too expressive model that risks overfitting the anomalous behaviors in the data. This non-trivial change improves NWR-GAE originally proposed for the unique purpose of dimension reduction now suitable for GAD tasks.

\begin{table}[!t]
\centering
\scalebox{0.8}{
\begin{tabular}{c|c|c|c}
\hline
\textbf{Approach}                                                                           & \textbf{\begin{tabular}[c]{@{}c@{}}Contextual\\ Anomaly\end{tabular}} & \textbf{\begin{tabular}[c]{@{}c@{}}Structural\\ Anomaly\end{tabular}} & \textbf{\begin{tabular}[c]{@{}c@{}}Joint-type\\ Anomaly\end{tabular}} \\ \hline
\begin{tabular}[c]{@{}c@{}} \textbf{Structure-based}\\ 
 SCAN~\cite{SCAN} and others~\cite{maulana2020centrality,shashikala2015outlier,tong2011non}\end{tabular} & $\times$               & $\checkmark$ & $\times$                                                                     \\ \hline
\begin{tabular}[c]{@{}c@{}}\textbf{Feature-based}\\ LOF~\cite{LOF}, IF~\cite{IF}, MLPAE~\cite{MLPAE}\end{tabular}            & $\checkmark$                                                                     & $\times$                                                                     & $\times$                                                                     \\ \hline
\begin{tabular}[c]{@{}c@{}}\textbf{GAE with proximity driven loss}\\AnomalyDAE~\cite{AnomalyDAE} , GCNAE~\cite{GCNAE} \\  DOMINANT~\cite{Dominant}\end{tabular}               & $\checkmark$                                                                     & $\checkmark$                                                                     & $\times$                                                                     \\ \hline
 \multirow{2}{*}{\textbf{GAD-NR (ours)}}                                                                       &  \multirow{2}{*}{$\checkmark$}                                                                &  \multirow{2}{*}{$\checkmark$}                                                                     &  \multirow{2}{*}{$\checkmark$}    \\ 
 & & &  \\
 \hline                                                             \end{tabular}}
 
\caption{\small SOTA methods perform well either on contextual or densely connected structural anomalies whereas \proj with its entire neighborhood reconstruction principle finds advantages for detecting both types of anomalies along with joint-type anomalies which are the nodes that connect a large number of nodes with different features.}
\vspace{-2em}
\label{tab:outlier_types}
\end{table}

We extensively compare \proj with state-of-the-art (SOTA) models on six real-world graph anomaly detection datasets that have been benchmarked recently~\cite{liu2022pygod}. 
\proj outperforms all baselines significantly (by up to 30\%$\uparrow$ in AUC) over five among these six datasets by following the settings in~\cite{liu2022pygod}. We also evaluate and demonstrate the capability of \proj on detecting each of the three types of anomalies. 

Note that in real-world applications, the types of anomalies are often unknown. The significance of \proj is that it allows detecting the real-world anomalies across different datasets (in~\cite{liu2022pygod}) with one fixed hyperparameter configuration, which illustrates the robustness of \proj. Further ablation studies also justify the effectiveness and computational efficiency of Gaussian approximation adopted by \proj for GAD when being compared with NWR-GAE~\cite{NWRGAE}.

The contributions of this paper can be summarized  as follows:
\begin{itemize}[leftmargin=4mm]
\item We designed a novel framework \proj for graph anomaly detection. \proj leverages the reconstruction loss of the entire neighborhood of a node from the node representation, which in principle can detect all three types of anomalies in Fig.~\ref{fig:outlier_types}. 
\item Technically, \proj adopts a Gaussian approximation of the distribution of neighbors' representations and computes a closed-form KL divergence as the reconstruction loss, which substantially improves the scalability and effectiveness of the approach.  
\item Extensive experiments on six real-world networks demonstrate the effectiveness of \proj compared to SOTA baselines, and the rationale of the design specifics of \proj.
\end{itemize}

\section{Related Works}

We put previous methods for GAD into three categories as follows.

\textbf{Structure-only-based methods:} Traditional graph anomaly detection focuses on detecting only structural anomalies. Many works in this category leverage spectral analysis of the adjacency matrix and its variants~\cite{spectralAD1,spectralAD2}. Recent methods define structural similarity measures for anomalies and then perform clustering approaches for detection~\cite{ANOMALOUS, SCAN}. Statistical features computed based on the graph structure such as in/out degrees, total weights of edges, number of neighbors of a node, or dense subgraphs can be utilized for GAD~\cite{akoglu2010oddball,ding2012intrusion,hooi2016fraudar}. 
However, these structure-based methods are only able to detect structural anomalies. They may detect some joint-type anomalies but they tend to make a 
 slot of false alarms as they miss the information from node attributes.

\textbf{Traditional methods for GAD over attributed networks:} In real-world applications, most of the graphs have node attributes (features). Nodes with inconsistent attributes have a high chance to be an anomaly node. Moreover, considering  the information on node attributes along with structure helps to locate anomalies more accurately.  
Detecting anomalies in attributed networks can be achieved by clustering methods~\cite{bojchevski2018bayesian,perozzi2014focused}, interaction with human experts~\cite{ding2019interactive}, group merging techniques~\cite{zhu2020mixedad}. Network embedding methods~\cite{perozzi2014deepwalk,grover2016node2vec,tang2015line} can also be applied to GAD on attributed graphs~\cite{bandyopadhyay2019outlier,bandyopadhyay2020outlier}.  
Network embeddings can be paired with anomaly detection techniques for tabular data such as density-based approaches~\cite{LOF}, and distance-based techniques~\cite{aggarwal2001outlier, IF} to find node anomalies on graphs. However, these approaches, since they process graph structure separately with node attributes, often fail to capture the synergy of graph structure and node attributes and may be suboptimal for GAD in some cases.

\textbf{Deep learning based GAD approaches:} Auto-Encoder framework that focuses on extracting principal components from the data via deep learning has been extensively applied in anomaly detection~\cite{DONE, GCNAE, Dominant, AnomalyDAE,luo2022comga}. Applying traditional autoencoders to node attributes~\cite{MLPAE} can only detect contextual anomalies.  
GAE built upon GNNs can combine node attributes and graph structure properly and can detect anomalies based on checking the reconstruction loss of node attributes or links~\cite{GCNAE, Dominant, AnomalyDAE}. But these works do not reconstruct the entire neighborhood for GAD. Rather, they use reconstruction error, and estimating Gaussian mixture density is also applied for GAD~\cite{li2019specae}. Some works view nodes with multiple views and a node may or may not be considered an anomaly in different views. These nodes hold attributes from multiple views of the identity. To capture such multi-view information, multiple GNNs are often applied~\cite{peng2020deep,sheng2019multi,wu2014boosting,wu2013multi,liu2022mul} for anomaly detection. GNNs have also been applied to detect anomalies in multiple scales~\cite{gutierrez2020multi}, and to detect anomalies and solve recommendation tasks simultaneouly~\cite{wang2019fdgars,zhang2020gcn}. More involved techniques such as self-supervised learning ~\cite{CONAD,jin2021anemone,liu2021anomaly,zhou2022improving,corsiniself,huang2023hop} and reinforcement learning~\cite{morales2021selective,ding2019interactive,langford2007epoch} have also been recently applied to GAD.
\vspace{-2mm}

\section{Notations and Problem Formulation}

In this work, we focus on detecting anomalous nodes over attributed static graphs.
An attributed graph $G  = (V, E, X) \in \mathcal{G}$ consists of a vertex set $V = \{1,2, \cdots, N\}$ 
and an edge set $E$. $X=[\cdots x_u^\top \cdots ]^\top\in \mathbb{R}^{|V| \times k}$ collect all node attributes and $x_u \in \mathbb{R}^k$ is the attribute for node $u$. The degree of node $u$ is denoted as $d_u$. 
This work focuses on unsupervised anomaly detection. Each node $u$ has an anomaly label $y_u$ where $y_u=0$ or $y_u=1$ implies node $u$ is normal or anomalous respectively. The goal is to design a detection method $f(G): \mathcal{G} \rightarrow \{0,1\}^N$ that associates each nodes with a label. However, these node labels are assumed to be unknown when designing $f$.   

Let $\mathcal{N}_u$ be the set of 1-hop neighbor nodes of node $u$. Let $\bar{\mathcal{N}}_u$ be an augmented set of 1-hop neighborhood of node $u$ that includes the attribute of node $u$, the set of the attributes of its neighbors, i.e., $\Bar{\mathcal{N}}_u \triangleq (x_u, \{x_v|v\in \mathcal{N}_u\})$. Our assumption to detect anomalous nodes is that given the label $y_u$, the distribution $\mathbb{P}(\bar{\mathcal{N}}_u|y_u)$ are different across norms and anomalies.  Here, we consider just one-hop neighborhood as a proof of concept, which is also often adequate for use cases in practice ~\cite{akoglu2012fast}.  The neighborhoods considered can be extended to the multi-hop case, while extra computation costs need to be paid in that scenario.

\section{Methodology}
In this section, we first provide the motivation of our method by narrating the potential drawbacks of previous graph auto-encoder methods. Then, we introduce \proj which is based on neighborhood reconstruction.

\subsection{Motivations}
AutoEncoder (AE) is an easy-to-use and effective framework for anomaly detection. The motivation of AE is to perform dimension reduction by compressing the high dimensional input data into a low dimensional latent representation~\cite{hinton2006reducing} via an encoder and reconstructing the original input with the help of a decoder. AE can be used for anomaly detection because such dimension reduction is expected to capture the principal properties of the data mostly corresponding to the normal data points. The data points that cannot be properly reconstructed via the decoder, i.e., with larger reconstruction losses tend to be anomalies.

Graph AutoEncoder (GAE) is used to perform dimension reduction of  graph data via a Graph Neural Network (GNN) as the encoder~\cite{kipf2016variational}. Given a graph $G=(V,E)$, GAE encodes graph data into node representations $\{h_v|v\in V\}$. The decoder of current GAE methods 
is to reconstruct the graph structure and node attributes from these node representations. Regarding graph-structure reconstruction, it typically relies on a mapping from the representations of two nodes to 0 or 1 that indicates whether there is an edge between them~\cite{Dominant, AnomalyDAE}, e.g., comparing $h_u^\top h_v$ with some threshold $\theta$ to reconstruct the edge. However, this procedure can only preserve proximity information of nodes in the graph, i.e., pushing node representations close if the corresponding nodes are directly connected in the graph, which may miss useful information for detecting anomalies. Moreover, by checking the reconstruction loss, one may only tell whether an edge is an anomaly. To detect node anomalies that are often more useful in practice, one needs to aggregate the reconstruction losses of edges into the node level, and how to properly aggregate these losses is not a trivial problem by itself and often depends on heuristics.

\subsection{GAE via Neighborhood Reconstruction}
\begin{figure}[!t]
\centering
    
    \includegraphics[width=0.48\textwidth]{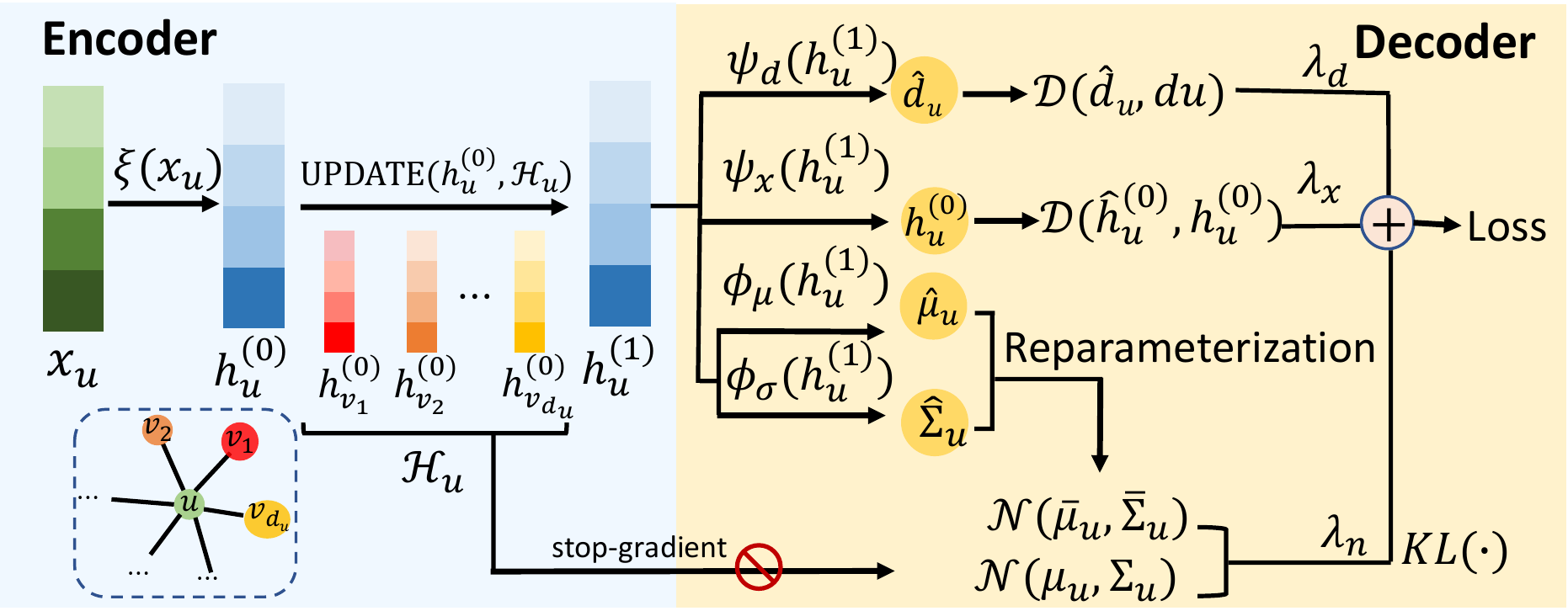}
    \caption{\small Model architecture of \proj. The encoder (left) performs dimension reduction with an MLP followed by a message passing GNN to obtain the hidden representation of a node. The decoder (right) reconstructs the self features and the node degree via MLPs and estimates the neighbor feature distribution with an MLP-predicted re-parameterized Gaussian distribution. Reconstructions of the self features and the node degree are optimized with MSE-loss whereas the KL-divergence between the ground truth and the learned neighbors' feature distribution is used for the optimization of the distribution estimation.}
    \label{fig:model_diagram}
\end{figure}

Our strategy to overcome the drawback of traditional GAEs lies in the first-principle idea of autoencoders. Autoencoders aim to perform dimension reduction of the data with the least loss to recover the original data. GAE encodes each node's attributes and the attributes of the nodes in its one-or-several-hop neighborhood into a node representation. Therefore, the node representation should be able to reconstruct the neighborhood and its attributes with the least loss. This idea leads to the design of GAE in this work. The model architecture is illustrated by Fig.~\ref{fig:model_diagram} and describes the pseudocode in Algorithm 1.

\subsubsection{The encoder}
The encoder $\Phi(\cdot)$ follows the common pipeline of message passing GNN~\cite{gilmer2017neural} e.g. GCN~\cite{GCN} or GraphSAGE~\cite{GRAPHSAGE}. A GNN will further iteratively aggregate the representations from the neighbors and combine them with one's own representation to update the representation. Specifically, let $h^{(0)}_u = x_u$. For $l=0,1,...,L-1$,
\begin{equation}
    h_u^{(l+1)}=\text{UPDATE}\,(h_u^{(l)}, \text{AGG}\,(\{ h_v^{(l)}:v \in \mathcal{N}_u\}), 
    \label{eq:GNN}
\end{equation}

The AGG function aggregates messages from the neighbors and the UPDATE function updates the node representations. Note that in practice, if the node attribute $x_u$ is extremely high dimensional and sparse, a random linear projection is used to encode them into a dense low-dimensional representation $h^{(0)}_u = \xi(x_u)$. 

\subsubsection{The decoder} Our decoder is designed based on the first principle of designing an autoencoder~\cite{hinton2006reducing}. We are supposed to reverse the procedure of Eq.~\eqref{eq:GNN} by using $h_u^{(L)}$ to reconstruct all the information within the $L$-hop neighborhood of $u$. In practice, it is computationally heavy for large $L$. In this work, we focus on reconstructing the information within just the one-hop neighborhood as a proof of concept, which we find is practically sufficient for GAD.

The information within just the one-hop neighborhood  $\bar{\mathcal{N}}_u$ consists of the attributes of the center node $h^{(0)}_u$ and the set of attributes of the direct neighbors of the center $\mathcal{H}_u=\{h^{(0)}_v|v\in \mathcal{N}_u\}$.

\textbf{Self reconstruction:} In order to reconstruct the attributes of the center node, we design a simple decoder that takes $h_u^{(L)}$ as input and reconstructs $h^{(0)}_u$ by a multi-layer perception (MLP) $\hat{h}_u^{(0)} = \psi_x(h_u^{(1)})$. 
Then, the self-reconstruction loss for node $u$ can be calculated as:
\begin{equation}\label{eq:att-loss}
    \mathcal{L}^x_u =  \mathcal{D}\left(h_u^{(0)}, \hat{h}_u^{(0)} \right)
\end{equation}
where $\mathcal{D}(\cdot, \cdot)$ is a distance function such as L2-distance that measures the discrepancy between the original attributes and the reconstructed attributes.

\textbf{Neighborhood reconstruction:} It is far from trivial to decode the set $\mathcal{H}_u$ from the compressed $h_u^{(L)}$. The difficulties come from two aspects. Firstly, the size of the set might vary across node $u\in V$. Secondly, the elements in the set do not have an order. Using an MLP to decode a set of a variable size from $h_u^{(L)}$ is impossible. 

Our idea is inspired by the recent work NWR-GAE~\cite{NWRGAE}. We regard the set $\mathcal{H}_u=\{h^{(0)}_v|v\in \mathcal{N}_u\}$ as $d_u$ many I.I.D. samples from a distribution $\mathbb{P}_u$. In fact, the empirical distribution of the elements in $\mathcal{H}_u$ is $\mathbb{P}_u^{\text{emp}}(h) = \frac{1}{d_u}\sum_{v\in\mathcal{N}_u} \delta(h-h^{(0)}_v)$ where $\delta(\cdot)$ is the dirac delta function. In this sense, we can decompose the neighborhood information into two parts, namely the number of neighbors (i.e. the node degree) $d_u$ and the distribution of neighbor's representations $\mathbb{P}_u$. The reconstruction procedure should reconstruct these two parts of information properly.

\begin{algorithm}[!t]
\label{alg:gadnr}
\caption{GAD-NR : Graph Anomaly Detection via Neighborhood Reconstruction}
\begin{algorithmic}[1]
\State {\textbf{Input:} Graph $G(V,E)$, Input Feature $X$, Anomaly Label $Y$} 
\State{\textbf{Encoder:}}
\For{$u \in V$}

\State $h^{(0)}_u = \xi(x_u)$
\State $h_u^{(1)}=\text{UPDATE}\,(h_u^{(0)}, \text{AGG}\,(\{ h_v^{(0)}:v \in \mathcal{N}_u\})$

\EndFor
\State{\textbf{Decoder:}}
\For{$u \in V$}
\State $\hat{h}_u^{(0)} = \psi_x(h_u^{(1)}), \hat{d}_u = \psi_d(h_u^{(1)})$ 

\State $\mu_{u}= \text{stop-gradient}(\frac{1}{d_u} \sum_{v \in \mathcal{N}_u} h_{v}^{(0)})$
\State $\Sigma_{u} = \text{stop-gradient}(\frac{1}{d_u-1} \sum_{v\in \mathcal{N}_u} (h_{v}^{(0)} - \mu_{u}) (h_{v}^{(0)} - \mu_{u})^\top$)
\State $\hat{\mu}_u = \phi_\mu (h^{(1)}_u), \hat{\Sigma}_u = \text{diag}(\exp(\phi_\sigma (h^{(1)}_u)))$

\For{$i = 1 \text{ to } q$} 
\Comment{Reparameterization}
\State{$\Bar{h}_i = \text{FNN}(z_i), z_i \sim \mathcal{N}(\Hat{\mu}_u,\Hat{\Sigma}_u)$}
\EndFor

\State $\Bar{\mu}_{u} = \frac{1}{q} \sum_{i=1}^{q} \Bar{h}_i, \Bar{\Sigma}_{u} = \frac{1}{d_u-1} \sum_{i=1}^{q} (\Bar{h}_i-\Bar{\mu}_u)(\Bar{h}_i-\Bar{\mu}_u)^\top$

\State \textbf{Weighted Loss Function}
\State  \scalebox{0.7}{ $\mathcal{L} = \lambda_x \sum\limits_{u \in V} \mathcal{D}\left(h_u^{(0)}, \hat{h}_u^{(0)} \right) + \lambda_d \sum\limits_{u \in V} \mathcal{D}\left(d_u, \hat{d}_u \right) +  \lambda_n \text{KL}(\mathcal{N}(\mu_{u},\boldsymbol{\Sigma}_u) || \mathcal{N}(\Bar{\mu}_{u} ,\Bar{\Sigma}_{u})) 
    $}
\EndFor

\end{algorithmic}
\end{algorithm}

\emph{Node degree reconstruction.} To reconstruct node degree $d_u$, we use another MLP that follows $\hat{d}_u = \psi_d(h_u^{(L)})$.

Then the node degree reconstruction loss for node $u$ is:
\begin{equation}\label{eq:degree-loss}
   \mathcal{L}^d_u  =  \mathcal{D}\left(d_u, \hat{d}_u \right)
\end{equation}
Here, we just use $\ell_2$-loss as the metric $\mathcal{D}$, though as node degrees are non-negative integers, we can also adopt discrete distributions such as Poisson distribution to model it.

\paragraph{Neighbors' representation distribution reconstruction.}

To reconstruct the distribution $\mathbb{P}_u$ from the node representation $h_u^{(L)}$, we first map $h_u^{(L)}$ to an estimation of the distribution $\hat{\mathbb{P}}_u$. As we do not know $\mathbb{P}_u$ in the population level, the first direct idea is to reconstruct the empirical distribution $\mathbb{P}_u^{\text{emp}}$ by following  NWR-GAE~\cite{NWRGAE}. The Wasserstein distance between $\hat{\mathbb{P}}_u$ and $\mathbb{P}_u^{\text{emp}}$ is adopted as the reconstruction loss in NWR-GAE~\cite{NWRGAE}. However, the computation of such a loss has a huge overhead, because it needs to solve a matching problem based on the Hungarian algorithm~\cite{HUNGARIAN}, which is of complexity $O(d_u^3)$. Moreover, empirically, we observe that reconstructing such an empirical distribution is likely to overfit the anomalies, which actually does harm to anomaly detection tasks. 

Therefore, we propose to reconstruct a multi-variate Gaussian approximation of $\mathbb{P}_u$. Specifically, given $\mathcal{H}_u=\{h^{(0)}_v|v\in \mathcal{N}_u\}$, we estimate the mean and covariance matrix of the neighbors' representations by following:

\begin{align}
\small
    \mu_{u}= \frac{1}{d_u} \sum_{v \in \mathcal{N}_u} h_{v}^{(0)}, \;\boldsymbol{\Sigma}_{u} = \frac{1}{d_u-1} \sum_{v\in \mathcal{N}_u} (h_{v}^{(0)} - \mu_{u}) (h_{v}^{(0)} - \mu_{u})^\top 
    \label{eq:meanVar1}
\end{align}
Then, we map $h_u^{(L)}$ to a  multi-variate Gaussian distribution $\hat{\mathbb{P}}_u$ through the following procedure. We sample $q$ neighborhood features $\Bar{h}_1, \cdots, \Bar{h}_q$ by transforming samples $z_1,\cdots,z_q$ from the distribution $\mathcal{N}(\Hat{\mu}_u,\Hat{\Sigma}_u)$ via a fully-connected neural network (FNN). Here the parameters $\Hat{\mu}_u,\Hat{\Sigma}_u$ are determined by \begin{equation}
     \hat{\mu}_u = \phi_\mu (h^{(L)}_u), \quad \boldsymbol{\hat{\Sigma}}_u = \text{diag}(\exp(\phi_\sigma (h^{(L)}_u))),
     \label{eq:meanVar_decode}
\end{equation}
where $\phi_\mu(\cdot)$ is an MLP, and each entry of $\phi_\sigma (h^{(L)}_u)$ is non-negative, which includes an MLP followed by $\exp(\cdot)$. Then, we estimate the mean and the covariance matrix of  reconstructed neighbors' features based on $\Bar{\mu}_{u} = \frac{1}{q} \sum_{i=1}^{q} \Bar{h}_i$ and $\Bar{\Sigma}_{u} = \frac{1}{d_u-1} \sum_{i=1}^{q} (\Bar{h}_i-\Bar{\mu}_u)(\Bar{h}_i-\Bar{\mu}_u)^\top$, respectively.

Given the two groups of parameters ($\mu_u, \Sigma_u$) and ($\Bar{\mu}_u,\Bar{\Sigma}_u$) for multi-variate Gaussian distributions, we adopt the KL divergence between these two distributions to measure the reconstruction loss:

\begin{equation}
\begin{aligned}
    \mathcal{L}_u^n = \text{KL}(\mathcal{N}(\mu_{u},\Sigma_u) || \mathcal{N}(\Bar{\mu}_{u} ,\Bar{\Sigma}_{u})) = \\
    \frac{1}{2} [ \log \frac{|\Sigma_u|}{|\Bar{\Sigma}_{u}|} - p 
    + \text{tr}(\Bar{\Sigma}_{u}^{-1}\Sigma_u) +(\mu_{u}-\Bar{\mu}_{u})^\top & \Bar{\Sigma}_{u}^{-1} (\mu_{u}-\Bar{\mu}_{u}).   \label{eq:n-loss}
    ]
\end{aligned}
\end{equation}
where $p$ is the dimension of  representation. Note that $\mu_{u}$ and $\boldsymbol{\Sigma}_{u}$ should not allow the gradient to pass through as they provide supervision signals. In practice, we may encounter the case when $d_u$ is smaller than $p$, which makes $\Sigma_u$ in Eq.~\eqref{eq:meanVar1} not full-ranked and causes a numerical problem. Therefore, we add an identity matrix to the covariance matrices, $\boldsymbol{\Sigma}_u\leftarrow \boldsymbol{\Sigma}_u+cI,\,\bar{\boldsymbol{\Sigma}}_{u}\leftarrow \bar{\boldsymbol{\Sigma}}_{u}+cI$ for some constant $c>0$ to compute Eq.~\eqref{eq:n-loss}. 

Note that the complexity of the above computation including Eqs.~\eqref{eq:meanVar1},\eqref{eq:meanVar_decode} and \eqref{eq:n-loss} is $O(d_u)$, 
 which significantly reduces the complexity of the pipeline in \cite{NWRGAE}.
 
 The remaining challenge is that since node degrees vary across different nodes, the computation of Eq.~\eqref{eq:meanVar1} is irregular. For this, we extend the package adopted in principal neighborhood aggregation~\cite{corso2020principal} to implement Eq.~\eqref{eq:meanVar1} efficiently in parallel across different nodes.

\subsubsection{The overall reconstruction loss.}
The overall reconstruction loss is a combination of the losses to reconstruct node self attributes in Eq.~\eqref{eq:att-loss}, node degrees in Eq.~\eqref{eq:degree-loss}, and neighbors' representation distributions in Eq.~\eqref{eq:n-loss}:

\begin{equation}\label{eq:loss}
    \mathcal{L} = \sum_{u \in V} \mathcal{L}_u',\;\text{where}\; \mathcal{L}_u'\triangleq \lambda_x \mathcal{L}_u^x + \lambda_d \mathcal{L}_u^d + \lambda_n  \mathcal{L}^n_u,
\end{equation}
and where $\lambda_x$, $\lambda_d$, and $\lambda_n$ are the hyper-parameters that control the weights of different types of reconstruction losses.

\subsection{Anomaly Detection}
We may adopt $\mathcal{L}_u$ in Eq.~\eqref{eq:loss} as the score to characterize how anomalous each node $u$ is to be. The greater score means the encoded information is harder to be reconstructed, and thus the corresponding node is more likely to be an anomaly. 
We may also adopt different hyperparameters $\lambda_x'$, $\lambda_d'$, and $\lambda_n'$ if we have different confidence or some prior knowledge about the type of anomaly to be detected. For example, if we tend to detect contextual anomalies, we can increase $\lambda_x'$. To encode such flexibility, we define the anomaly score $\hat{y}_u$ in the following general form 
\begin{equation} \label{eq:obj}
    \hat{y}_u = \mathcal{L}_u'(\lambda_x',\lambda_d',\lambda_n') = \lambda_x' \mathcal{L}_u^x + \lambda_d' \mathcal{L}_u^d + \lambda_n'  \mathcal{L}^n_u,
\end{equation}
ranking which tells the nodes that are more likely to be anomalies. Although different weights here emphasize the detection of different types of anomalies, in Sec.~\ref{sec:exp}, we show that \proj is robust to the selection of these weights, where a fixed choice of these weights is sufficient to outperform baselines to detect real-world anomalies across different datasets.

\subsection{Improvements over NWR-GAE}

Here, we would like to provide a more direct explanation how \proj advances the idea of neighborhood reconstruction (previously proposed in NWR-GAE~\cite{tang2022rethinking} for the purpose of dimension reduction instead of GAD) to better fit GAD tasks. NWR-GAE is built upon an optimal transport loss and needs to run a complicated Hungarian matching algorithm~\cite{HUNGARIAN} for each node to reconstruct its neighbors' attributes to compute the loss function. Such complexity is $O(d^3)$ for a node of degree $d$. \proj regards the representations of neighbors' attributes as samples from a Gaussian distribution and adopts KL divergence~\cite{KLdivergence} between Gaussian distributions as the reconstruction loss, which has a closed form and has complexity $O(d)$.
This approximation is crucial for GAD tasks: 

NWR-GAE did not adopt such approximation because NWR-GAE aims to perform dimension reduction. Achieving a low reduction error is the ultimate goal of NWR-GAE. Therefore, NWR-GAE should be sufficiently expressive to make low-dimensional representations recover high-dimensional data. However, GAD tasks have different goals. A model for GAD should not be too expressive, and otherwise risks overfitting the anomalies. \proj just adopts the correct trade-off, where Gaussian approximation (by just checking the first and second moments of the distributions) is adopted, which not only improves anomaly detection accuracy but also substantially reduces computational complexity. Moreover, NWR-GAE also supports reconstructing multi-hop neighbors. However, we found that multi-hop reconstruction did not get obvious improvement in the task of GAD while introducing much computational overhead, so \proj only considers the first hop in practice.

\section{Experiment} \label{sec:exp}
In this section, we extensively compare \proj with several baseline methods for graph anomaly detection. Specifically, we aim to answer the following questions:
\begin{itemize}[leftmargin=4mm]
\item How does neighborhood reconstruction facilitate in the  performance improvement of \proj for GAD?
\item Which part of the \proj is important for different types of anomaly detection?
\item How do important hyperparameters such as the size of hidden representations, and the weights before different types of reconstruction losses affect the performance of \proj?
\item How does the adopted Gaussian approximations of neighborhood feature distributions improve the running time efficiency of \proj?
\end{itemize}

\subsection{Datasets and Baselines}
We incorporate six real-world datasets (Cora, Weibo, Reddit, Disney, Books, and Enron) and fourteen baseline anomaly detection models for our comparison following the BOND~\cite{liu2022bond} paper.  Among the baseline models, we included feature-based  models LOF~\cite{LOF}, IF~\cite{IF}, MLPAE~\cite{MLPAE} and structure-based AD models, SCAN~\cite{SCAN}. Also, we performed comparisons of \proj with models that focus on both structures and attributes via residual reconstruction error Radar~\cite{Radar} and ANOMALOUS~\cite{ANOMALOUS}. Lastly, we incorporated some popular generative models for GAD, which include autoencoder architecture GCNAE~\cite{GCNAE}, DOMINANT~\cite{Dominant}, DONE and AdONE~\cite{DONE}, AnomalyDAE~\cite{AnomalyDAE}, adversarial learning-based method GAAN~\cite{GAAN} and also contrastive learning-based methods CONAD~\cite{CONAD}. 
\vspace{-1em}

\subsection{Experimental Settings}
Our first experimental setting follows the benchmark paper BOND~\cite{liu2022pygod}. Note that among the datasets, Weibo, Reddit, Disney, Books, and Enron have real-world anomaly labels. For the Cora dataset, there are no real benchmark anomaly labels, so we follow the benchmark paper BOND where the union of contextual and structural anomalies are considered anomaly labels for evaluation in the Cora dataset. The results are reported in Table~\ref{tab:real_world_ground_truth}. We call this setting the benchmark anomaly detection.  Moreover, we attempt to study the performance to detect each type of anomalies separately, so for each dataset including those with real-world labels,  we also inject contextual, structural, and joint-type anomalies for evaluation, which give the later results in Table~\ref{tab:injected_anomaly_detection}. Due to the page limitation, we present contextual and merged the structural and joint-type anomaly detection results together in this work.

Contextual anomalies are nodes whose attributes are significantly different from their neighboring nodes. Hence, to generate this type of anomaly for a target node $u$, its feature $x_u$ is replaced with another randomly sampled node $v$'s feature $x_v$ that has the largest Euclidean distance with $x_u$. Let $n$ denote the number of contextual anomaly nodes and $q$ denote the number of candidate nodes randomly sampled in the above procedure. Structural anomalies are nodes that are densely connected in contrast to sparsely connected normal nodes. To inject structural outliers, we consider $m$ nodes at random and then make them fully connected and this process will be repeated for $n$  times to generate $n$ such cliques of size $m$. Following the BOND paper, we approximately set $q$ and $m$ as twice the avg. degree for most datasets. To add joint-type anomalies in different datasets, we choose $n$ nodes randomly as anomalies. Then, we connect each of these $n$ nodes with randomly sampled $m$ other nodes. Therefore, those anomalous nodes can be treated as nodes with high degrees and connected to neighbors with different types of features. We utilized the PyGOD library~\cite{liu2022pygod} for the injection of contextual and structural anomalies and for running the baseline anomaly detection models.

\begin{table*}[!t]
\centering
\scalebox{0.69}{
\begin{tabular}{c|c|c|c|c|c|c}
\hline
\textbf{Algorithm}  & \textbf{Cora}             & \textbf{Weibo}            & \textbf{Reddit}          & \textbf{Disney}          & \textbf{Books}           & \textbf{Enron}            \\ \hline
LOF        & 69.9 ± 0.0 (69.9)  & 56.5 ± 0.0 (56.5)  & \underline{57.2 ± 0.0 (57.2)} & 47.9 ± 0.0 (47.9) & 36.5 ± 0.0 (36.5) & 46.4 ± 0.0 (46.4)  \\ 
IF         & 64.4 ± 1.5 (67.4)  & 53.5 ± 2.8 (57.5)  & 45.2 ± 1.7 (47.5) & 57.6 ± 2.9 (63.1) & 43.0 ± 1.8 (47.5) & 40.1 ± 1.4 (43.1)  \\ 
MLPAE      & 70.9 ± 0.0 (70.9)  & 82.1 ± 3.6 (86.1)  & 50.6 ± 0.0 (50.6) & 49.2 ± 5.7 (64.1) & 42.5 ± 5.6 (52.6) & 73.1 ± 0.0 (73.1)  \\ 
SCAN       & 62.8 ± 4.5 (72.6)  & 63.7 ± 5.6 (70.8)  & 49.9 ± 0.3 (50.0) & 50.5 ± 4.0 (56.1) & 49.8 ± 1.7 (52.4) & 52.8 ± 3.4 (58.1)  \\ 
Radar      & 65.0 ± 1.3 (66.0)  & \textbf{98.9 ± 0.1 (99.0)}  & 54.9 ± 1.2 (56.9) & 51.8 ± 0.0 (51.8) & 52.8 ± 0.0 (52.8) & \underline{80.8 ± 0.0 (80.8)}  \\ 
ANOMALOUS  & 55.0 ± 10.3 (68.0) & \textbf{98.9 ± 0.1 (99.0)}  & 54.9 ± 5.6 (60.4) & 51.8 ± 0.0 (51.8) & 52.8 ± 0.0 (52.8) & \underline{80.8 ± 0.0 (80.8)}  \\ 
GCNAE      & 70.9 ± 0.0 (70.9)  & 90.8 ± 1.2 (92.5)  & 50.6 ± 0.0 (50.6) & 42.2 ± 7.9 (52.7) & 50.0 ± 4.5 (57.9) & 66.6 ± 7.8 (80.1)  \\ 
DOMINANT   & 82.7 ± 5.6 (84.3)  & 85.0 ± 14.6 (92.5) & 56.0 ± 0.2 (56.4) & 47.1 ± 4.5 (54.9) & 50.1 ± 5.0 (58.1) & 73.1 ± 8.9 (85.0)  \\ 
DONE       & 82.4 ± 5.6 (87.9)  & 85.3 ± 4.1 (88.7)  & 53.9 ± 2.9 (59.7) & 41.7 ± 6.2 (50.6) & 43.2 ± 4.0 (52.6) & 46.7 ± 6.1 (67.1)  \\ 
AdONE      & 81.5 ± 4.5 (87.4)  & 84.6 ± 2.2 (87.6)  & 50.4 ± 4.5 (58.1) & 48.8 ± 5.1 (59.2) & 53.6 ± 2.0 (56.1) & 44.5 ± 2.9 (53.6)  \\ 
AnomalyDAE & 83.4 ± 2.3 (85.3)  & 91.5 ± 1.2 (92.8)  & 55.7 ± 0.4 (56.3) & 48.8 ± 2.2 (55.4) & 62.2 ± 8.1 (73.2) & 54.3 ± 11.2 (69.1) \\ 
GAAN       & 74.2 ± 0.9 (76.1)  & \underline{92.5 ± 0.0 (92.5)}  & 55.4 ± 0.4 (56.0) & 48.0 ± 0.0 (48.0) & 54.9 ± 5.0 (61.9) & 73.1 ± 0.0 (73.1)  \\ 
GUIDE      & 74.7 ± 1.3 (77.5)  & OOM\_C           & OOM\_C          & 38.8 ± 8.9 (52.5) & 48.4 ± 4.6(63.5)  & OOM\_C           \\ 
CONAD      & 78.8 ± 9.6 (84.3)  & 85.4 ± 14.3 (92.7) & 56.1 ± 0.1 (56.4) & 48.0 ± 3.5 (53.1) & 52.2 ± 6.9 (62.9) & 71.9 ± 4.9 (84.9)  \\ \hline 
 \proj (w/o feat. recon.)    & \underline{83.41 ± 2.18 (85.41)} &
69.64 ± 0.75  (70.44) &
50.00 ± 0.44 (50.88) &
\underline{74.11 ± 0.18 (76.17)} &62.32 ± 3.41 (65.43) &
70.20 ± 1.16 (71.72)
\\
 \proj (w/o degree recon.)            & 
82.25 ± 1.37 (83.04) & 
70.09 ± 1.20 (71.50) &
49.01 ± 0.29 (50.05) &
76.25 ± 0.37 (79.09) &
\underline{64.08 ± 3.13 (68.94)} &
72.44 ± 1.33 (75.81)
\\
 \proj (w/o neighbor recon.) &
76.47 ± 3.57 (80.47) & 
69.10 ± 1.10 (70.25)&
48.67 ± 2.04 (50.67)& 
60.69 ± 1.24 (63.69)& 
49.46 ± 2.09 (52.46)& 
56.01 ± 3.00 (59.01)\\ 

 \textbf{\proj}           & 
\textbf{87.55 ± 2.56 (88.40)}&
87.71 ± 5.39 (92.09)&
\textbf{57.99 ± 1.67 (59.90)}&
\textbf{76.76 ± 2.75 (80.03)}&
\textbf{65.71 ± 4.98 (69.79)}&
\textbf{80.87 ± 2.95 (82.92)}
\\ \hline      
\end{tabular}}
\caption{ \small Performance comparison (ROC-AUC) of \proj in benchmark anomaly detection for six different real-world datasets (injected anomaly for Cora dataset). For the results of baseline methods, we followed the BOND~\cite{liu2022bond} paper where the \textit{avg performance ± the STD of perf. (max perf.)} is reported. For our model~\proj, we fix hyperparameters ~$\lambda_x = 0.8$, $\lambda_d = 0.5$ and $\lambda_n = 0.001$ and report the \textit{avg performance ± the STD of perf.} for all datasets including the best  performance in each dataset with tuned hyperparameters. The best and second best performances are mentioned in \textbf{bold} and \underline{underlined} respectively and \textbf{$OOM\_C$} indicates out of memory with regard to GPU. }
\vspace{-2mm}
\label{tab:real_world_ground_truth}\
\end{table*}

\begin{table*}[t]
\centering
\scalebox{0.5}{
\begin{tabular}{c|cccccc} \hline 
\textbf{Algorithm}           & \textbf{Cora}         & \textbf{Weibo}        & \textbf{Reddit}       & \textbf{Disney}       & \textbf{Books}        & \textbf{Enron}        \\ \hline
MLPAE                        & {\ul 88.90 ± 0.00}    & 90.61 ± 0.02          & 51.91 ± 4.55          & 86.36 ± 0.00          & 53.00 ± 13.99         & 68.74 ± 16.08         \\
SCAN                         & 49.80 ± 0.50          & 48.46 ± 0.00          & 48.59 ± 0.00          & 62.81 ± 0.00          & 50.15 ± 0.00          & 40.41 ± 0.00          \\
Radar                        & 50.20 ± 0.60          & 72.31 ± 0.00          & 49.98 ± 0.00          & 79.89 ± 0.00          & {\ul 66.30 ± 0.00}    & {\ul 79.51 ± 0.00}    \\
ANOMALOUS                    & 51.10 ± 1.30          & 72.31 ± 0.00          & 49.65 ± 1.26          & 79.89 ± 0.00          & {\ul 66.30 ± 0.00}    & 48.92 ± 0.85          \\
GCNAE                        & {\ul 88.90 ± 0.00}    & \textbf{90.79 ± 0.35} & 52.02 ± 0.36          & {\ul 87.52 ± 2.32}    & 40.77 ± 1.72          & 59.52 ± 15.23         \\
DOMINANT                     & 71.90 ± 6.60          & 57.07 ± 0.34          & 47.96 ± 0.39          & 65.62 ± 9.53          & 50.13 ± 5.33          & 63.84 ± 0.27          \\
DONE                        & 70.2 ± 8.30         &     81.75 ± 1.00      &  46.52 ± 1.47   &   69.26 ± 5.33       &     38.52 ± 2.08     & 59.90 ± 6.31\\
AdONE                        & 73.90 ± 5.00          & 83.68 ± 0.47          & 46.61 ± 3.42          & 86.12 ± 1.83          & 65.00 ± 2.63          & 62.19 ± 5.37          \\
AnomalyDAE                   & 80.20 ± 2.80          & 79.40 ± 3.67          & 49.33 ± 2.28          & 81.02 ± 8.13          & 40.43 ± 18.56         & 68.12 ± 14.93         \\
GAAN                         & 88.70 ±0.10           & {\ul 90.68 ± 0.14}    & 49.64 ± 1.12          & 86.39 ± 0.13          & 32.99 ± 22.38         & 44.78 ± 14.20         \\
GUIDE                        & 88.30 ± 0.80          & OOM\_C                & OOM\_C                & 78.40 ± 0.62          & 57.59 ± 0.08          & OOM\_C                \\
CONAD                        & 72.50 ± 5.80          & 56.62 ± 0.37          & 47.70 ± 0.08          & 59.67 ± 7.00          & 54.04 ± 7.22          & 71.23 ± 0.54          \\ \hline
GAD-NR (w/o feat. recon.)    & 58.52 ± 2.32          & 57.66 ± 3.22          & 49.60 ± 2.11          & 82.98 ± 2.60          & 62.11 ± 5.01          & 68.27 ± 2.09          \\
GAD-NR (w/o degree recon.)   & 73.04 ± 2.60          & 64.28 ± 6.35          & {\ul 54.10 ± 1.96}    & \textbf{91.10 ± 3.56} & 60.07 ± 1.17          & 75.25 ± 3.57          \\
GAD-NR (w/o neighbor recon.) & 71.52 ± 3.59          & 61.45 ± 4.42          & 53.45 ± 1.15          & 57.11 ± 2.50          & 55.36 ± 3.19          & 68.23 ± 3.34          \\
GAD-NR                       & \textbf{89.10 ± 3.10} & 87.53 ± 3.54          & \textbf{55.15 ± 2.41} & 85.72 ± 1.31          & \textbf{74.73 ± 1.50} & \textbf{85.79 ± 2.65} \\ \hline
\end{tabular}}
\quad
\scalebox{0.5}{
\begin{tabular}{c|ccccccc} \hline 
 \textbf{Algorithm}           & \textbf{Cora}         & \textbf{Weibo}        & \textbf{Reddit}       & \textbf{Disney}       & \textbf{Books}        & \textbf{Enron}        \\ \hline
 MLPAE                        & 51.28 ± 0.43          & 50.42 ± 0.00          & 50.10 ± 0.52          & 58.77 ± 0.00          & 52.52 ± 2.48          & 48.30 ± 1.21          \\
 SCAN                         &\underline{82.35 ± 0.00}          & 49.86 ± 0.00          & \textbf{98.11 ± 0.00} & 64.82 ± 0.00          & 62.21 ± 0.00          & 48.08 ± 0.00          \\
 Radar                        & 62.37 ± 0.00          & \underline{ 60.04 ± 0.00}    & 65.70 ± 0.00          & 63.77 ± 0.00          & 32.69 ± 0.00          & 53.29 ± 0.00          \\
 
 ANOMALOUS                    & 45.39 ± 1.15          & \underline{ 60.04 ± 0.00}    & 58.08 ± 29.77         & 63.77 ± 0.00          & 32.69 ± 0.00          & 52.81 ± 0.23          \\
 
 GCNAE                        & 51.19 ± 0.00          & 50.66 ± 0.06          & 51.30 ± 0.43          & 52.82 ± 0.90          & 31.93 ± 0.35          & 47.33 ± 6.84          \\
 
 DOMINANT                     & 77.59 ± 0.03          & 49.01 ± 0.92          & 93.18 ± 0.00          & 34.84 ± 4.38          & 63.99 ± 0.11          & 62.70 ± 0.10          \\
 
 DONE                         & 72.34 ± 14.02         & 55.56 ± 0.90          & 65.61 ± 10.50         & 71.12 ± 1.40          & \underline{ 87.86 ± 0.42}    & 50.88 ± 2.51          \\
 
 AdONE                        & 81.32 ± 1.39          & 59.23 ± 0.53          & 80.07 ± 2.26          & 71.43 ± 2.76          & \textbf{88.30 ± 0.61} & 55.93 ± 2.05          \\
 AnomalyDAE                   & 80.73 ± 0.93          & 49.16 ± 1.33          & 47.03 ± 2.66          & 49.93 ± 6.26          & 32.25 ± 9.01          & 49.83 ± 1.89          \\
 GAAN                         & 53.26 ± 0.84          & 53.09 ± 0.18          & 62.78 ± 4.85          & 60.20 ± 0.35          & 78.98 ± 1.00          & 59.88 ± 2.35          \\
 GUIDE                        & 52.03 ± 2.36          & OOM\_C                & OOM\_C                & 55.20 ± 1.50          & 63.72 ± 0.89          & OOM\_C                \\
 CONAD                        & 78.85 ± 0.02          & 55.48 ± 0.36          & \underline{93.25 ± 0.17}          & 34.49 ± 2.32          & 63.97 ± 0.31          & 59.98 ± 0.45          \\  
 \hline
   GAD-NR (w/o feat loss)       & 
 73.23 ± 0.61 &
 50.81 ± 0.28 &
 91.95 ± 0.03 &
 \underline{73.23 ± 1.61} &
 80.06 ± 5.49 &
 \underline{79.18 ± 2.01} \\
 GAD-NR (w/o degree loss)     &
 74.28 ± 1.67 &
 53.33 ± 2.31 &
 87.69 ± 6.47 &
 66.47 ± 0.51 &
 82.95 ± 0.30 &
 72.19 ± 0.06 \\
 GAD-NR (w/o neigh loss)      &
 67.51 ± 0.04 &
 50.29 ± 1.11 &
 91.34 ± 0.11 &
 60.54 ± 2.06 &
 56.28 ± 3.03 &
 75.65 ± 1.39 \\
 GAD-NR                       &  \textbf{83.55 ± 3.03} &
 \textbf{62.35 ± 1.05} &
 92.01 ± 0.73 &
 \textbf{74.81 ± 4.39} &
 85.01 ± 7.90 &
 \textbf{82.22 ± 2.14}
 \\ \hline
\end{tabular}}
\caption{\small Performance comparison (ROC-AUC) of \proj in contextual (left) and structural + joint-type (right) anomaly detection for different real-world datasets. The best and second best performances are mentioned in \textbf{bold} and \underline{underlined} respectively and \textbf{$OOM\_C$} indicates out of memory with regard to GPU.}
\label{tab:injected_anomaly_detection}
\end{table*}

\begin{table*}[!ht]
\centering
\scalebox{0.65}{
\begin{tabular}{c|c|c|c|c|c|c|c}
\hline
\begin{tabular}[c]{@{}c@{}}Comparison\\ Type\end{tabular}                         & Algorithm   & Cora         & Weibo        & Reddit       & Disney       & Books        & Enron        \\ \hline
\multirow{2}{*}{\begin{tabular}[c]{@{}c@{}}Performance\\ Comparison\end{tabular}} & NWR-GAE      & 84.28 ± 0.06 & 73.68 ± 3.13 & 51.20 ± 1.16 & 75.56 ± 1.65 & 79.75 ± 2.48 & 80.24 ± 2.43 \\ 
                                                                                  & GAD-NR      & \textbf{87.55 ± 2.56} & \textbf{89.71 ± 2.39} & \textbf{57.99 ± 1.67} & \textbf{76.76 ± 2.75} & \textbf{80.64 ± 4.98} & \textbf{85.77 ± 3.86} \\ \hline
\multirow{2}{*}{\begin{tabular}[c]{@{}c@{}}Running time \\ Seconds per epoch\end{tabular}}                                                    & NWR-GAE      & 51.270       & 48.312       & 55.379       & 0.603        & 7.288        & 65.152       \\  
                                                                                  & GAD-NR      & \textbf{2.35}         & \textbf{0.544}        & \textbf{0.095}        & \textbf{0.035}        & \textbf{0.0874}       & \textbf{0.0109}       \\ 
                                                                                  \hline
\end{tabular}}
 \caption{\small Direct Performance Comparison between NWR-GAE~\cite{NWRGAE} and our model \proj}
 \vspace{-2mm}
\label{tab:nwrgaevsgadnr}
\end{table*}

\begin{figure*}[t]
  \centering
  \includegraphics[trim={0cm 0.3cm 0cm 0cm},clip, width=0.3\textwidth]{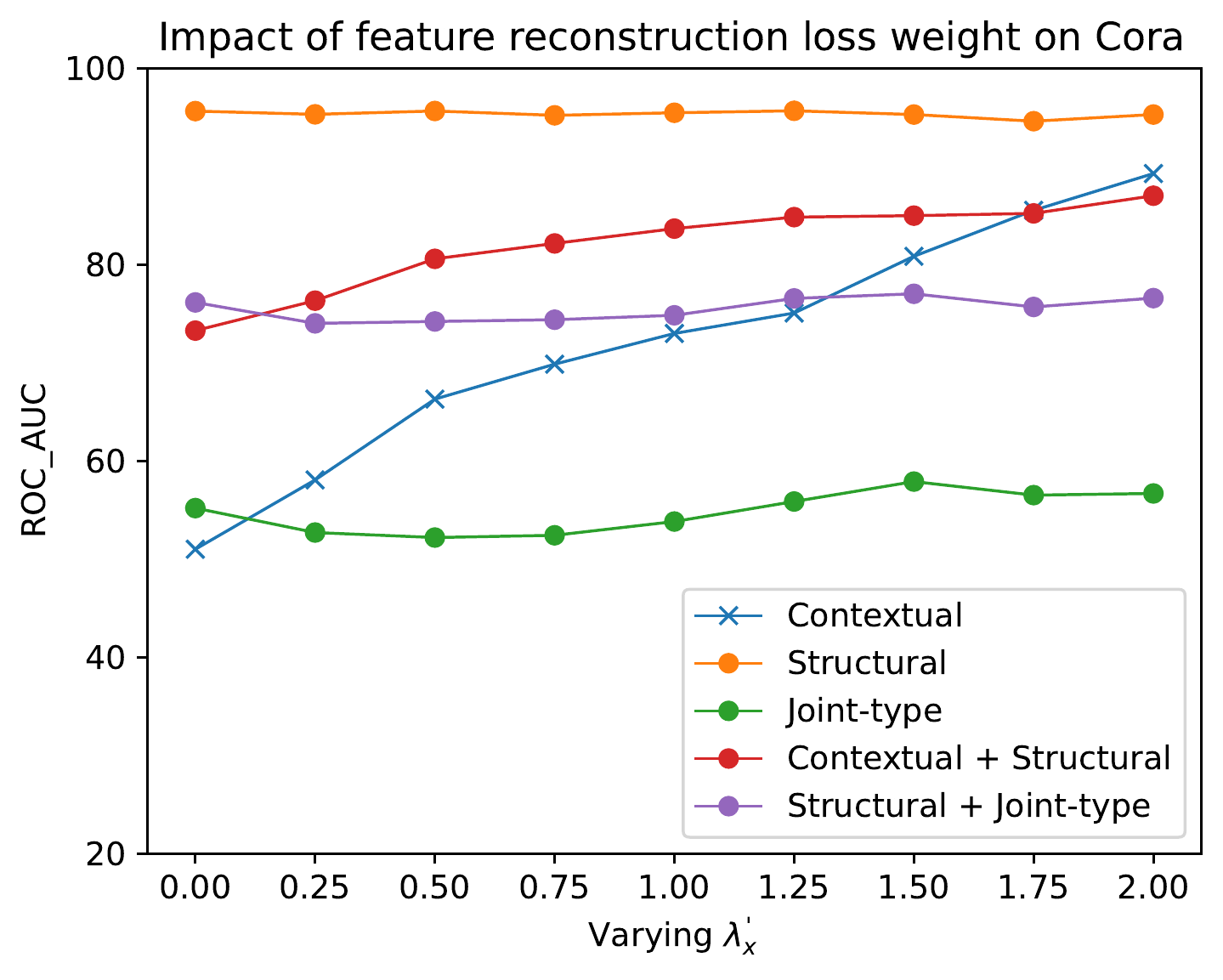}
  \includegraphics[trim={0cm 0.3cm 0cm 0cm},clip, width=0.3\textwidth]{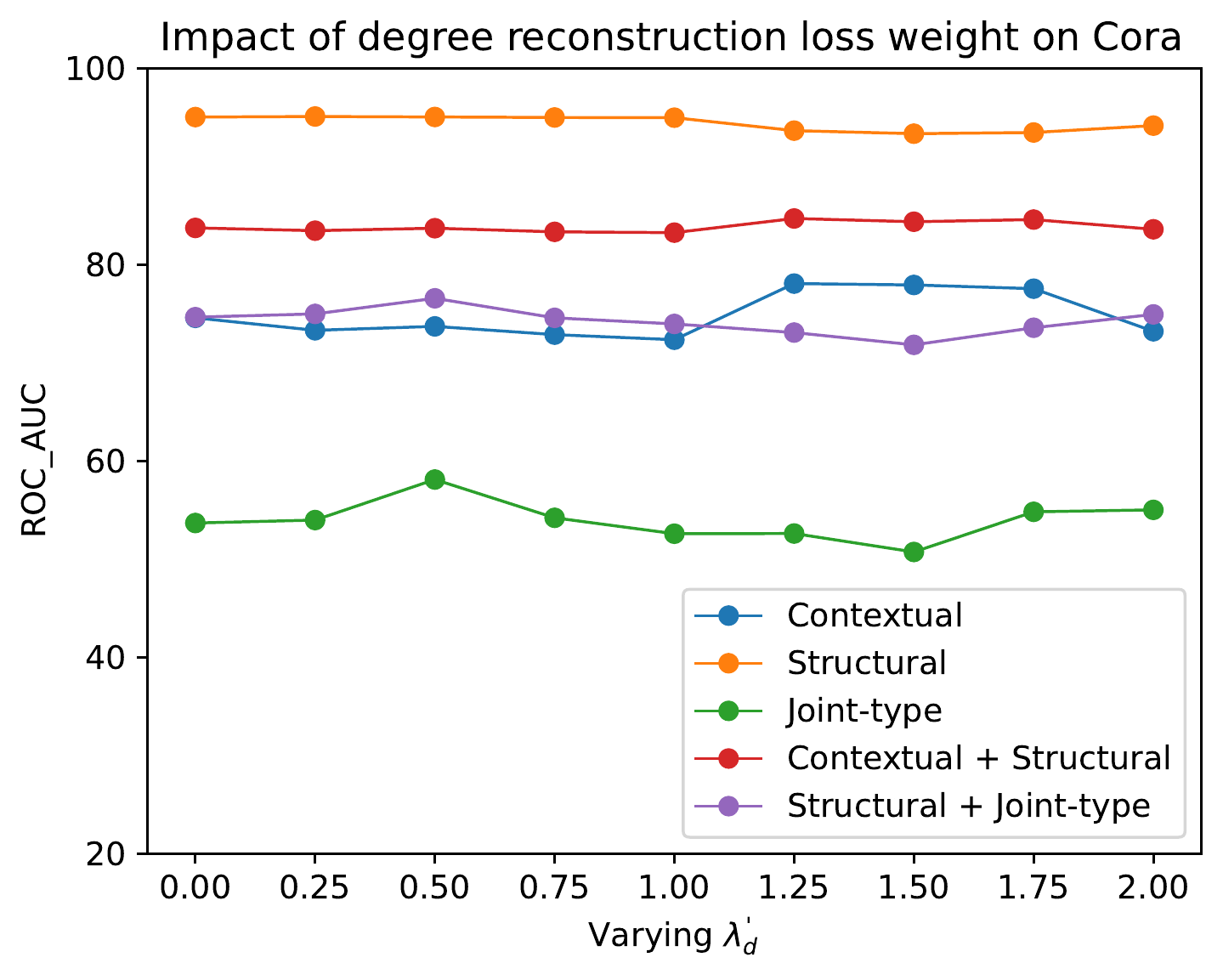}
  \includegraphics[trim={0cm 0.3cm 0cm 0cm},clip, width=0.3\textwidth]{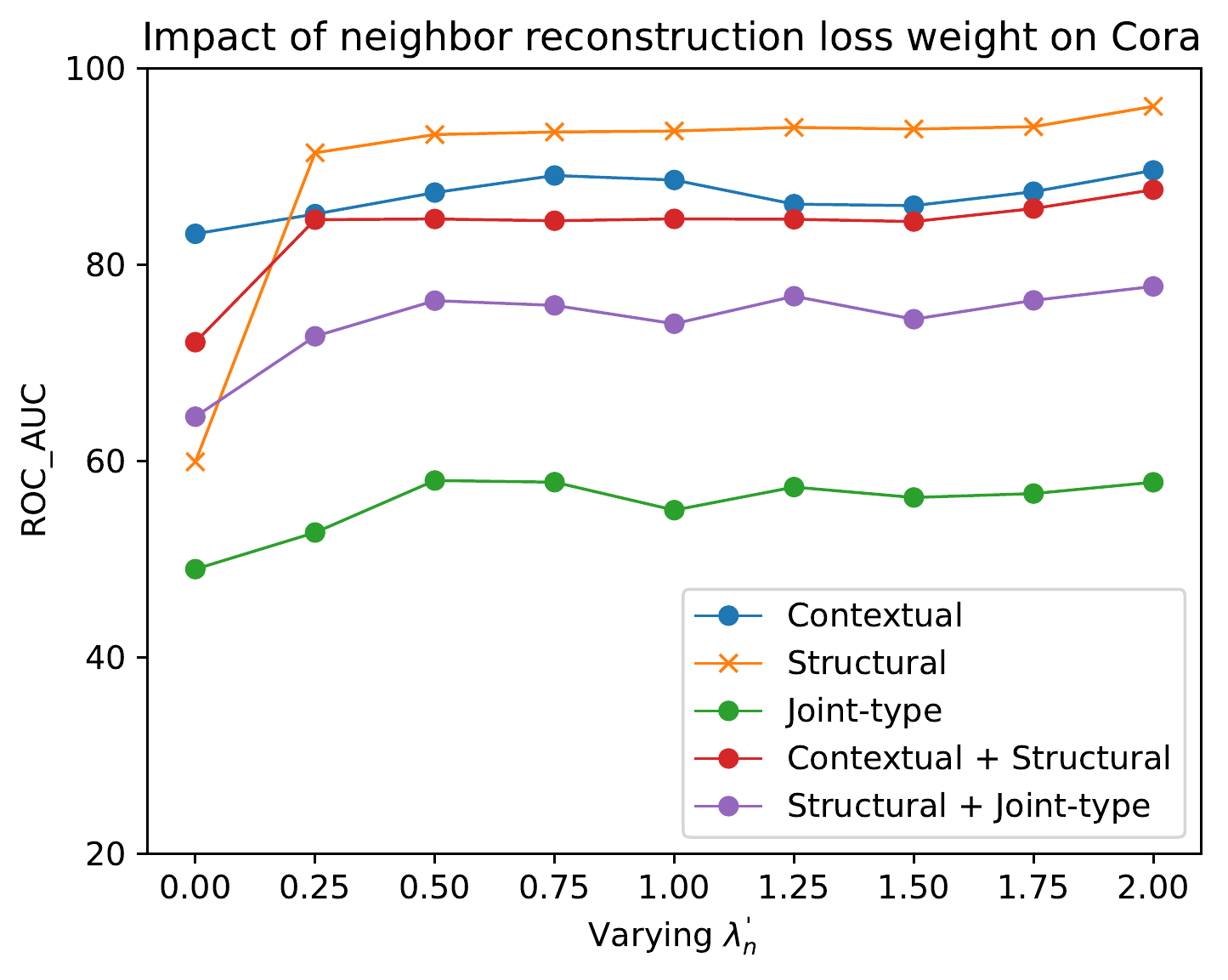}
 \includegraphics[trim={0cm 0.3cm 0cm 0cm},clip, width=0.3\textwidth]{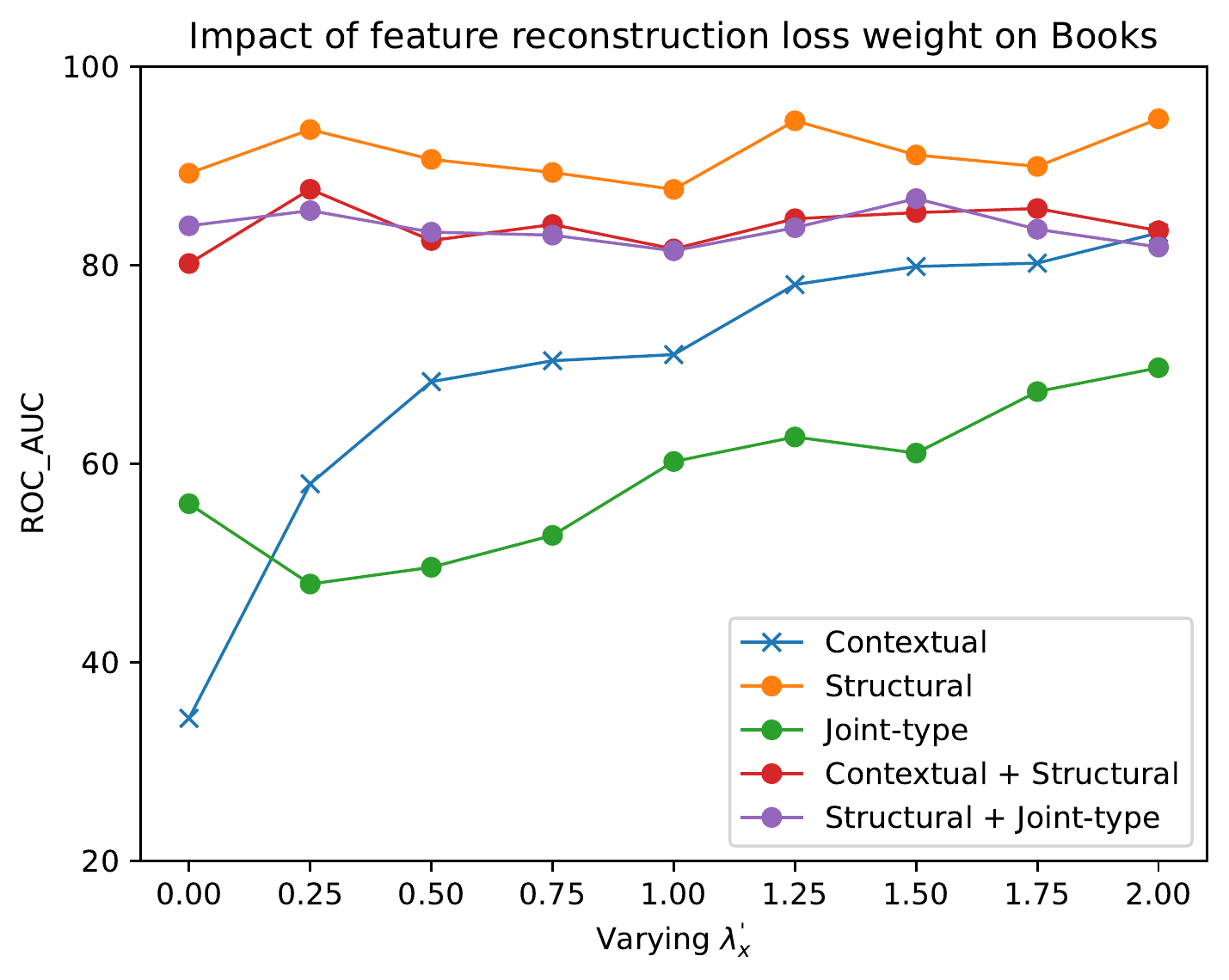}
 \includegraphics[trim={0cm 0.3cm 0cm 0cm},clip, width=0.3\textwidth]{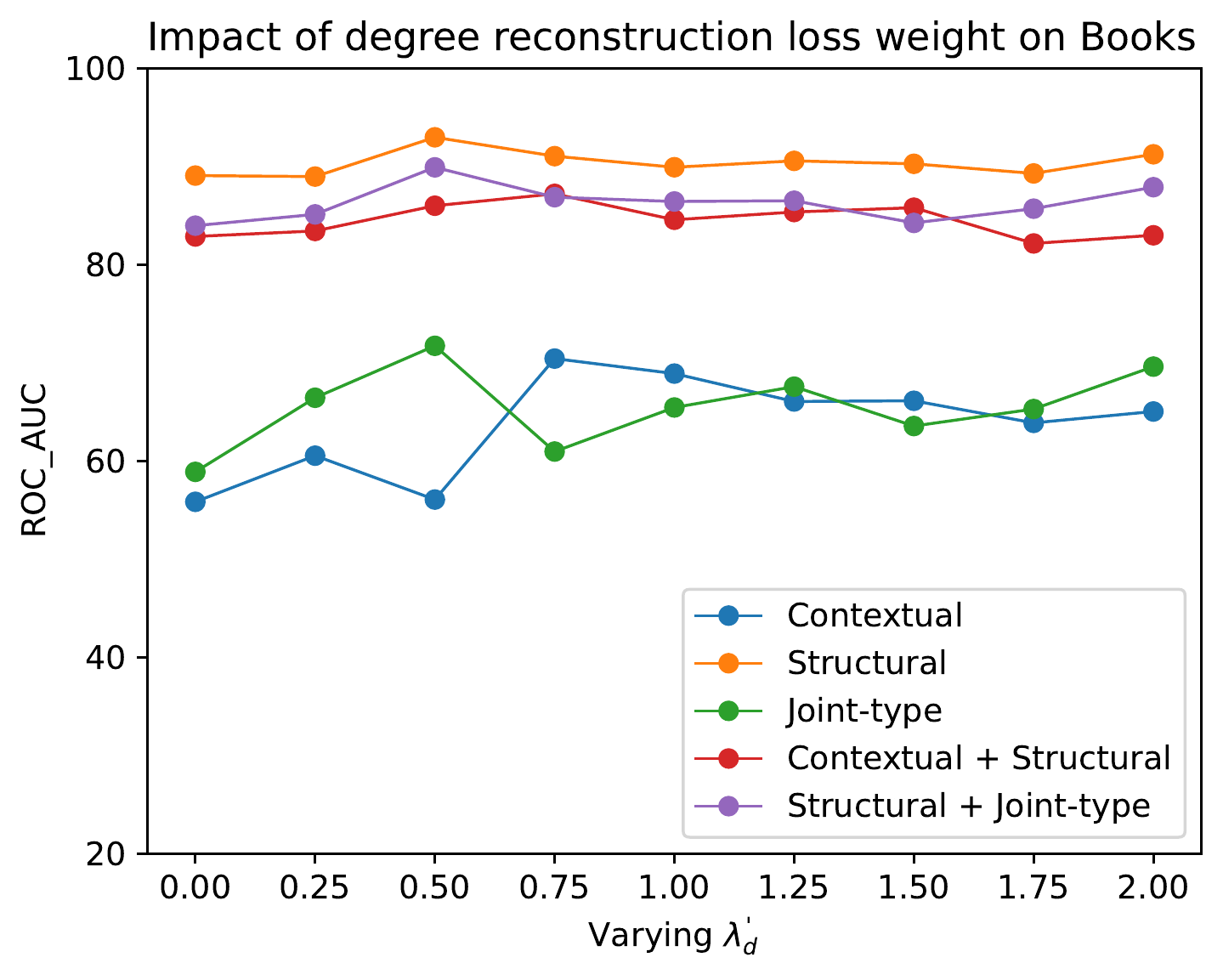}
 \includegraphics[trim={0cm 0.3cm 0cm 0cm},clip, width=0.3\textwidth]{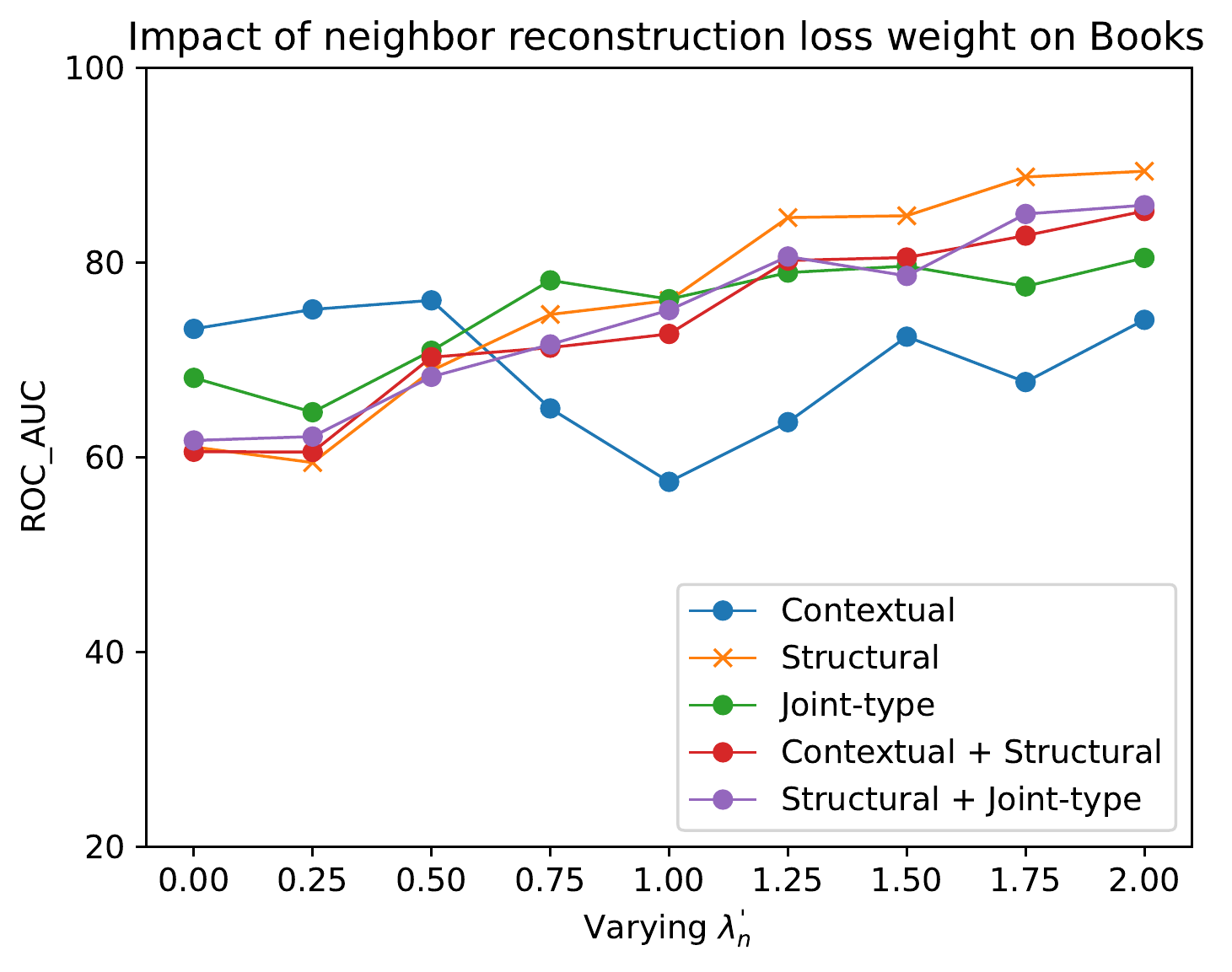}
  \caption{ \small Impacts of varying feature reconstruction weight loss $\lambda_x'$, degree reconstruction weight loss $\lambda_d'$ and neighbor reconstruction weight loss $\lambda_n'$ in Eq.~\eqref{eq:obj} on detecting different types of anomalies in the Cora (top) and Books (bottom) dataset.}
  \vspace{-1em}
\label{fig:ablation_study_lambda}
\end{figure*}

\textbf{Hyperparameter Tuning:} In practice, we often do not know the anomaly labels to tune the parameters. Typically, the way to choose hyperparameters is based on some expert experiences and a good model should be robust by using such a set of hyperparameters. Hence, we fix \proj's encoder as GCN with hidden dimension 16 (expect for Cora, where 128 is used) and fix the hyperparameters of the decoder as $\lambda_x=0.8$, $\lambda_d=0.5$, and $\lambda_n=0.001$ to run the experiments for all datasets five times and report the averaged performance with standard deviation. We compare \proj's performance obtained via this fixed hyperparameter setting with the averaged performance of baselines proposed in the benchmark work~\cite{liu2022bond}. Our experiments demonstrate that our model \proj can outperform the baselines by setting a set of hyperparameters that are not sensitive to the datasets, which makes the most practical sense.

To compare with the best performance of baselines reported in~\cite{liu2022bond}, we also performed a grid search of the hyperparameters of \proj for each dataset as follows: 1) Self-attribute reconstruction weight, $\lambda_x \in \{0.1, 0.5, 0.8, 0.9\}$ and $\lambda_x' \in \{0.25, 0.5, 0.75, 1.0, 1.25, 1.5,$ $ 1.75, 2.0\}$, 2) Degree reconstruction weight, $\lambda_d \in \{0.1, 0.5, 0.8\}$ and $\lambda_d' \in \{0.25, 0.5, 0.75, 1.0, 1.25, 1.5,$ $ 1.75, 2.0\}$, 3) Neighborhood reconstruction weight $\lambda_n \in \{0.001, 0.5,$ $0.8\}$ and $\lambda_n' \in$ \{0.25, 0.5, 0.75,$ $1.0, 1.25, 1.5, 1.75, 2.0\}, 4) The number of dimension of the hidden layer, $d \in \{ 8, 16, 32, 64, 128\}$, 5) Encoder GNN, $ \Phi \in \{ \text{GCN, GraphSAGE, GIN} \}$.

\textbf{Hardware:} All the experiments are performed on a Linux server with a 2.99GHz AMD EPYC 7313 16-Core Processor and 1 NVIDIA A10 GPU with 24GB memory.

\subsection{Evaluation Metric}
We adopt the area under the ROC curve as the evaluation metric. The ROC curve is created by plotting the true positive rate against the false positive rate at various threshold settings. In the experiment, we regard the anomaly nodes as positive classes and compute AUC for it. AUC equals 1 means that the model makes a perfect prediction, and AUC equals 0.5 means that the model has no distinguishing ability. AUC is better than accuracy when evaluating the anomaly detection task since it is not sensitive to the imbalanced class distribution of the data.

\subsection{Detection Performance Comparison}

\subsubsection{\proj shows the superior performance in different types of anomaly detection.} 
In Table~\ref{tab:real_world_ground_truth}, we show the results of \proj on the benchmark anomaly detection with baseline models.  
In Table~\ref{tab:injected_anomaly_detection}, we present the results in injected contextual, structural and joint-type anomaly detection. From the results, we can observe that \proj outperforms the baseline methods
over most of the datasets in detecting benchmark anomaly labels, contextual anomaly labels, and structural + joint-type anomalies.

The key reason behind the performance improvement can be attributed to the entire neighborhood reconstruction around a target node, which includes its self-feature reconstruction, degree reconstruction, and neighbor-feature distribution reconstruction. 

The feature-based models like MLPAE perform quite well at detecting contextual anomalies, specifically in the Cora dataset as they put emphasis on self-feature reconstruction. However, MLPAE performs worse to detect joint-type and structural anomalies as they ignore the graph structure. 
Approaches that only consider the structure information, for example, SCAN performs exceedingly well in detecting structural + joint-type anomalies but they are incapable of doing well in contextual anomaly detection. In the case of the GAE-based models, the performance is more competitive while still worse than \proj  in Table ~\ref{tab:real_world_ground_truth} and Table \ref{tab:injected_anomaly_detection}.

\subsubsection{Impacts of different types of reconstruction losses.} 
In Table~\ref{tab:real_world_ground_truth}, Table~\ref{tab:injected_anomaly_detection} along with the performance of \proj with all three types of reconstruction loss, we have also shown results by removing each part from the loss function of the \proj's decoder, i.e., setting $\lambda_x=0$, $\lambda_d=0$ or $\lambda_n=0$ in Eq.~\eqref{eq:loss}. From the results in Table~\ref{tab:injected_anomaly_detection}, it is clearly visible that without the neighborhood reconstruction part $\lambda_n=0$, the performance of \proj drops the most in both types of anomaly detection. 

From the results of Table ~\ref{tab:injected_anomaly_detection}, we can observe that without the self-feature reconstruction loss ($\lambda_x=0$), the performance of \proj drops down heavily when detecting contextual anomalies. When detecting the structural + joint-type anomalies, the performance decay of \proj is moderate, which matches the expectation. By removing the loss for degree reconstruction ($\lambda_d=0$), \proj also suffers from some performance decay. However, such decay is less severe compared to that of removing self-feature reconstruction for detecting contextual anomalies or that of removing neighbors' feature distribution reconstruction for detecting structural anomalies and joint-type anomalies.

\subsubsection{Performance Comparison with NWR-GAE~\cite{NWRGAE}}  We compare  \proj with NWR-GAE~\cite{NWRGAE} in two aspects: performance and running time.
For performance, we are adding the benchmark anomaly detection performance comparison between NWR-GAE and our model \proj in Table~\ref{tab:nwrgaevsgadnr}. We can observe that the performance of \proj is significantly better than NWR-GAE in all six datasets. 
NWR-GAE directly tries to match the empirical distribution of neighbor representation, by which NWR-GAE may capture neighbors' features more precisely (also time consuming) but it tends to overfit anomalous behaviors, compared to the Gaussian approximation that \proj adopts. For running time comparison, we also added the comparison  between NWR-GAE and our model GAD-NR in Table~\ref{tab:nwrgaevsgadnr}. Optimizing the KL-divergence leads to a running time complexity of $O(d)$ from the neighborhood matching the Hungarian algorithm's running time complexity of $O(d^3)$. Therefore, \proj is far more scalable on a relatively large graph dataset compared with NWR-GAE for detecting anomalies.

\vspace{-1em}

\subsection{Hyperparameter Analysis}

\subsubsection{Impacts of tuning $\lambda_x',\,\lambda_d',\,\text{and}\,\lambda_n'$} We present the trend of \proj's performance on different types of anomaly detection by varying the weights in Eq.~\eqref{eq:obj} to perform detection in Figure ~\ref{fig:ablation_study_lambda}. While increasing the weight for self-feature reconstruction  $\lambda_x'$ in Figure ~\ref{fig:ablation_study_lambda} left top, we have observed the performance curve of contextual anomaly detection (blue) is very steep. Similar in ~\ref{fig:ablation_study_lambda} left bottom, a growing trend has been observed on both contextual and joint-type anomaly detection performance curve (blue and green). The reason is intuitive. With higher weights for self-feature reconstruction, the decoder of \proj tends to assign higher importance to contextual anomalies as well as joint-type anomalies.  By varying the weight for degree reconstruction, $\lambda_d'$ in Figure ~\ref{fig:ablation_study_lambda} middle column, the change in performance is not that significant across different types of anomalies. This is because contextual and structural anomalies do not have much change in node degrees. For the joint-type anomalies, where node degrees may provide useful signals for detecting, only checking node degrees is often insufficient to determine an anomaly. This is because a normal node can also have higher degrees. Node degree reconstruction should be paired with neighbors' feature distribution reconstruction together to provide effective anomaly detection. 
Lastly, in Figure ~\ref{fig:ablation_study_lambda} right column, when we vary the weight  $\lambda_n'$ for neighborhood reconstruction, we notice a significant performance gain in joint-type anomaly detection and structural + joint-type anomaly detection, which demonstrates the effectiveness of neighborhood reconstruction via leveraging signals from their neighborhoods.

\begin{table}[h]
\scalebox{0.6}{
\begin{tabular}{c|cccc|cccc}
\hline
\textbf{Dataset}    & \multicolumn{4}{c|}{\textbf{Cora}}                                                                                               & \multicolumn{4}{c}{\textbf{Reddit}}                                                                                             \\ \hline
\textbf{\diagbox[width=10em]{Models}{\# Dimensions}} & \multicolumn{1}{c|}{32}             & \multicolumn{1}{c|}{64}             & \multicolumn{1}{c|}{128}            & 256            & \multicolumn{1}{c|}{8}              & \multicolumn{1}{c|}{16}             & \multicolumn{1}{c|}{32}             & 64             \\ \hline
MLPAE               & \multicolumn{1}{c|}{71.07}          & \multicolumn{1}{c|}{71.07}          & \multicolumn{1}{c|}{71.08}          & 70.51          & \multicolumn{1}{c|}{47.57}          & \multicolumn{1}{c|}{52.12}          & \multicolumn{1}{c|}{51.79}          & 51.92          \\ \hline
GCNAE               & \multicolumn{1}{c|}{71.47}          & \multicolumn{1}{c|}{71.52}          & \multicolumn{1}{c|}{71.63}          & 70.84          & \multicolumn{1}{c|}{50.88}          & \multicolumn{1}{c|}{51.29}          & \multicolumn{1}{c|}{51.81}          & 52.10          \\ \hline
DOMINANT            & \multicolumn{1}{c|}{84.52}          & \multicolumn{1}{c|}{84.77}          & \multicolumn{1}{c|}{84.90}          & 76.77          & \multicolumn{1}{c|}{52.93}          & \multicolumn{1}{c|}{52.94}          & \multicolumn{1}{c|}{52.99}          & 53.04          \\ \hline
DONE                & \multicolumn{1}{c|}{84.19}          & \multicolumn{1}{c|}{84.29}          & \multicolumn{1}{c|}{86.52}          & 78.29          & \multicolumn{1}{c|}{52.39}          & \multicolumn{1}{c|}{52.41}          & \multicolumn{1}{c|}{55.15}          & 55.86          \\ \hline
AdONE               & \multicolumn{1}{c|}{84.01}          & \multicolumn{1}{c|}{84.43}          & \multicolumn{1}{c|}{84.87}          & 73.42          & \multicolumn{1}{c|}{57.78}          & \multicolumn{1}{c|}{53.64}          & \multicolumn{1}{c|}{54.85}          & 55.14          \\ \hline
GAAN                & \multicolumn{1}{c|}{74.33}          & \multicolumn{1}{c|}{74.15}          & \multicolumn{1}{c|}{74.32}          & 76.15          & \multicolumn{1}{c|}{50.21}          & \multicolumn{1}{c|}{52.32}          & \multicolumn{1}{c|}{52.42}          & 52.79          \\ \hline
CONAD               & \multicolumn{1}{c|}{84.54}          & \multicolumn{1}{c|}{84.79}          & \multicolumn{1}{c|}{84.46}          & 76.15          & \multicolumn{1}{c|}{52.74}          & \multicolumn{1}{c|}{52.95}          & \multicolumn{1}{c|}{53.03}          & 53.13          \\ \hline
\textbf{GAD-NR}     & \multicolumn{1}{c|}{\textbf{86.38}} & \multicolumn{1}{c|}{\textbf{86.93}} & \multicolumn{1}{c|}{\textbf{87.55}} & \textbf{82.67} & \multicolumn{1}{c|}{\textbf{53.12}} & \multicolumn{1}{c|}{\textbf{57.99}} & \multicolumn{1}{c|}{\textbf{58.12}} & \textbf{56.07} \\ \hline
\end{tabular}}
\caption{\small Performance comparison of \proj with different latent dimension sizes for detecting benchmark anomalies in Cora and Reddit datasets.}
\vspace{-2em}
\label{tab:dimension_size_experiment}
\end{table}

\subsubsection{Impact of the latent representation's dimension.}

In Table~\ref{tab:dimension_size_experiment}, we present the performance of \proj on benchmark anomaly detection in the Cora and Reddit datasets by varying the dimension size of hidden representations. From the results, we can observe that the performance of \proj gradually increases as the latent dimension increases for Cora (32 to 128) and Reddit (8 to 32) compared to other GAE-based methods. We think using neighborhood reconstruction is the reason behind the gradual performance improvement of \proj. Other autoencoders can only increase the capability of latent representations by increasing the latent dimension. Instead, \proj can also increase the supervision strength of neighborhood reconstruction by increasing the latent dimension. When the dimension size increases even more e.g. 256 in Cora and 64 in Reddit, the anomaly detection performance of \proj  drops. With a higher latent dimension size, the model becomes too much expressive and it can overfit the anomalies. For anomaly detection, we are expected to capture normal behaviors instead of making models memorize all information in the data, especially abnormal behaviors. Therefore, we need to strike a balance between model expressiveness and the proportion of normal information extracted for the best anomaly detection performance.

\section{Conclusion}

In this study, we introduce \proj, for identifying anomalous nodes in graph structures. \proj is based on a graph auto-encoder that reconstructs the neighborhood information from node representations generated by a GNN encoder. The reconstruction process encompasses a self-feature representation, degree reconstruction, and the distribution of neighboring node representations, thus allowing the detection of various anomalies including contextual, structural, and joint-type anomalies. Experimental results on six real-world datasets demonstrate the effectiveness of neighborhood reconstruction in identifying different types of anomalies. \proj outperforms state-of-the-art GAD baselines in five out of the six datasets in benchmark evaluations. Additionally, \proj provides flexibility and potential for detecting different types of anomalies through the combination of different types of reconstruction loss with varying weights. \proj also shows the robustness of the selection of weights to detect real-world anomalies.

\section{Acknowledgement}
AR and PL were supported partially by the Sony Award and the NSF grant IIS-2239565. The authors would like to greatly thank Prof. Bruno Ribeiro, Prof. Sharon (Yixuan) Li, and anonymous reviewers for their insightful suggestions to improve the paper.

\bibliographystyle{plain}
\bibliography{ref}

\section{Appendix}

\begin{table*}[t]
\centering
\scalebox{0.7}{
\begin{tabular}{c|cccc|c|c|c|c|c|c} \hline

\multicolumn{1}{l|}{\multirow{2}{*}{\textbf{Dataset}}} & \multicolumn{1}{l}{\multirow{2}{*}{\textbf{\# Nodes}}} & \multicolumn{1}{l}{\multirow{2}{*}{\textbf{\# Edges}}} & \multicolumn{1}{l}{\multirow{2}{*}{\textbf{\# Feat.}}} & \multirow{2}{*}{\textbf{Avg. Degree}} & \multicolumn{2}{c|}{\textbf{\# Contextual Anomaly}} & \multicolumn{2}{c}{\textbf{\# Joint/Structural Anomalies}} & \multicolumn{1}{|l}{\multirow{2}{*}{\textbf{\centering  \begin{tabular}[c]{@{}l@{}}\# Anomalies\\    (Combined)\end{tabular}}}} & \multicolumn{1}{|l}{\multirow{2}{*}{\textbf{Ratio}}} \\  \cline{6-9}
\multicolumn{1}{l|}{}                                  & \multicolumn{1}{l}{}                                   & \multicolumn{1}{l}{}                                   & \multicolumn{1}{l}{}                                   &                                       &     \textbf{n}               & \textbf{k}              & \textbf{n}               & \textbf{m}              & \multicolumn{1}{l}{}                                                                                           & \multicolumn{1}{|l}{}                                \\ \hline
Cora~\cite{Cora}                                                  & 2,708                                                  & 11,060                                                 & 1,433                                                  & 4.1                                   & 70                       & 10                      & 70                       & 10                      & 138                                                                                                            & 5.1\%                                               \\
Weibo~\cite{Weibo}                                                 & 8,405                                                  & 407,963                                                & 400                                           & 48.5                                  & 434                      & 10                      & 434                      & 10                      & 868                                                                                                            & 10.3\%                                              \\
Reddit~\cite{reddit1,reddit2}                                                & 10,984                                       & 168,016                                                & 64                                                     & 15.3                                  & 183                      & 30                      & 183                      & 30                      & 366                                                                                                            & 3.3\%                                               \\
Disney~\cite{disney}                                                & 124                                                    & 335                                                    & 28                                            & 2.7                          & 3               & 5                       & 3                        & 5                       & 6                                                                                                              & 4.8\%                                               \\
Books~\cite{booksandenron}                                                 & 1,418                                         & 3,695                                         & 21                                            & 2.6                          & 14              & 5                       & 14                       & 5                       & 28                                                                                                             & 2.0\%                                               \\
Enron~\cite{booksandenron}                                                 & 13,533                                                 & 176,987                                                & 18                                                     & 13.1                                  & 3                        & 25                      & 3                        & 25                      & 6                                                                                                              & 0.4\%   \\ \hline                                            
\end{tabular}}
\caption{ \small Statistics of graph anomaly detection datasets. For contextual, structural and joint-type anomaly injection, we follow the setting in BOND~\cite{liu2022bond} paper regarding the anomaly ratio and the parameters $k$ and $m$ are twice the avg. degree approximately.}
\label{tab:dataset_description}
\end{table*}

\begin{table*}[!h]
\centering
\scalebox{0.7}{
\begin{tabular}{c|cccccc} \hline 
 \textbf{Algorithm}           & \textbf{Cora}         & \textbf{Weibo}        & \textbf{Reddit}       & \textbf{Disney}       & \textbf{Books}        & \textbf{Enron}        \\ \hline
 MLPAE                        & 45.49 ± 1.19          & 48.56 ± 0.00          & 49.95 ± 2.37          & 56.06 ± 0.00          & 47.16 ± 10.69         & 51.20 ± 11.92         \\
 SCAN                         & 45.37 ± 0.00          & 49.66 ± 0.00          & 50.26 ± 0.00          & 47.38 ± 0.00          & 49.49 ± 0.00          & 57.10 ± 0.00          \\
 Radar                        & 48.78 ± 0.00          & 49.86 ± 0.00          & \underline{54.13 ± 0.00}          & 47.93 ± 0.00          & 57.84 ± 0.00          & 36.36 ± 0.00          \\
 ANOMALOUS                    & 45.02 ± 0.77          & 49.86 ± 0.00          & 50.65 ± 2.00          & 47.93 ± 0.00          & 57.84 ± 0.00          & 35.30 ± 0.64          \\
 GCNAE                        & 45.06 ± 0.01          & 48.85 ± 0.12          & 51.00 ± 0.28          & 57.41 ± 1.19          & 46.47 ± 2.15          & 27.32 ± 2.89          \\
 DOMINANT                     & 48.89 ± 0.03          & 48.97 ± 0.32          & 51.75 ± 0.03          & 34.10 ± 10.06         & 51.89 ± 1.96          & 42.99 ± 0.64          \\
 DONE                         & 52.12 ± 2.12          & 49.13 ± 0.19          & 50.33 ± 1.27          & 55.76 ± 0.88          & 39.81 ± 2.61          & 42.05 ± 10.62         \\
AdONE                        & \underline{53.47 ± 1.12}          & 49.83 ± 1.13          & 49.92 ± 1.87          & 63.25 ± 3.10          & 41.77 ± 3.10          & 51.13 ± 15.66         \\
 AnomalyDAE                   & 45.47 ± 0.55          & \underline{49.95 ± 1.01}          & 49.20 ± 1.44          & 51.79 ± 5.21          & 51.88 ± 3.77          & 32.21 ± 16.59         \\
 GAAN                         & 49.23 ± 1.53          & 49.03 ± 0.13          & 48.65 ± 2.48          & 55.76 ± 0.28          & 46.18 ± 3.91          & 46.33 ± 5.38          \\
 GUIDE                        & 51.20 ± 1.21          & OOM\_C                & OOM\_C                & 55.48 ± 0.41          & 45.74 ± 0.68          & OOM\_C                \\
 CONAD                        & 50.30 ± 0.03          & 49.78 ± 0.44          & 51.76 ± 0.02          & 42.09 ± 9.80          & 50.89 ± 2.34          & \underline{58.47 ± 0.13}          \\  \hline
  GAD-NR (w/o feat loss)       &
 52.14 ± 0.43 &
 49.36 ± 0.37 &
 52.97 ± 0.74 &
 55.14 ± 2.56 &
 \underline{58.13 ± 1.37} &
 54.89 ± 0.84 \\
 GAD-NR (w/o degree loss)     &
 45.30 ± 0.65 &
 47.63 ± 0.55 &
 52.15 ± 0.71 &
 56.17 ± 0.21 &
 53.72 ± 3.17 &
 52.63 ± 0.38

          \\
           GAD-NR (w/o neigh loss)      &
 48.72 ± 0.09 &
 48.71 ± 1.43 &
 51.45 ± 1.43 &
 \underline{69.20 ± 3.71} &
 56.30 ± 0.84 &
 51.44 ± 0.23 \\
 GAD-NR &
 \textbf{57.43 ± 2.04} &
 \textbf{52.57 ± 0.46} &
 \textbf{54.72 ± 0.71} &
 \textbf{74.50 ± 2.16} &
 \textbf{67.04 ± 4.16} &
 \textbf{80.34 ± 8.27}
 \\
 \hline
\end{tabular}}
\caption{ \small Performance comparison (ROC-AUC) of \proj in joint-type anomaly detection for different real-world datasets. The best and second best performances are mentioned in \textbf{bold} and \underline{underlined} respectively and \textbf{$OOM\_C$} indicates out of memory with regard to GPU.}
\vspace{-0.5em}
\label{tab:outlier_type_comparison}
\end{table*}

 \subsection{Baseline Description}
 \label{sec:baseline_description}

\begin{itemize}[leftmargin=4mm]
    \item \textbf{LOF:} \cite{LOF} Local Outlier Factor (LOF) computes how isolated a node is compared to its neighborhood which only uses the node feature and the neighborhood is computed using k-nearest neighbors.

    \item \textbf{IF:} \cite{IF} Isolation Forest is an ensemble method of base trees for anomaly detection where the closeness of individual instances to the root of the tree is used to define the decision boundary.

    \item \textbf{MLPAE:} \cite{MLPAE} MLPAE leverages multi-layer perceptron as the encoder and decoder to reconstruct the node features and reconstruction loss of node feature is used as the anomaly score of a node.

    \item \textbf{SCAN:} \cite{SCAN} Structural Clustering Algorithm for Networks (SCAN) takes only the structure of the graph to detect clusters that are regarded as the structural anomalies of the graph.

    \item \textbf{Radar:} \cite{Radar} Radar takes both structure and node attribute information as the input and detects the anomaly nodes which are different from the majority in terms of attribute residual and network coherence and residual reconstruction norm is considered as anomaly score.

    \item \textbf{GCNAE:} \cite{GCNAE} GCNAE employs the encoder-decoder architecture where the encoder is a GCN that uses node attribute and feature information to find a latent representation and the decoder employs two GCNs to reconstruct node attributes and graph structure. The reconstruction error of the decoder is used as the anomaly score.
    \item \textbf{DOMINANT:} \cite{Dominant} DOMINANT also follows the encoder-decoder architecture the encoder is a two-layer GCN while the node attribute decoder is a two-layer GCN and the adjacency information is decoded using the dot product. Anomaly score is defined as the combination of both decoders.  
    \item \textbf{DONE:} \cite{DONE} DONE utilizes MLPs for the encoder and decoder architecture. The node embeddings and anomaly scores are simultaneously optimized with a unified loss function.
    \item \textbf{AnomalyDAE:} \cite{AnomalyDAE} AnomalyDAE utilizes a structure AE and attribute AE to detect anomaly nodes where the structure AE takes both adjacency matrix and node attribute as the input and attribute decoder utilizes structure and attribute embeddings to reconstruct node attribute. 
    \item \textbf{GAAN:} \cite{GAAN} GAN-based architecture used for anomaly detection method in GAAN which generates fake graphs using MLP and  encodes graph information using another MLP. The discriminator is trained to decide whether a graph is fake or real. The real node detection confidence and node attribute reconstruction are used as the anomaly score.

    \item \textbf{GUIDE:} \cite{GUIDE} GUIDE preprocess the structure information where node motif degree is used to represent structure  vector and the rest of the architecture is the same as DONE.

    \item \textbf{CONAD:} \cite{CONAD} CONAD generates augmented graphs to impose prior knowledge of anomaly nodes and contrastive learning loss is used for optimization.

    \item \textbf{NWRGAE:}~\cite{NWRGAE} Neighborhood Reconstruction-based Graph Auto Encoder performs neighborhood matching with optimal transport loss along with self-feature  and degree reconstruction  for node classification and role identification tasks.
\end{itemize}

\subsection{Dataset Description}

\label{sec:datset_description}
\begin{itemize}[leftmargin=4mm]
\item \textbf{Cora}~\cite{Cora}: Cora dataset represents a citation network where the nodes are the research papers and edges represent the citation information between the papers. Node attributes of this dataset represent the bag-of-words vectors from the paper’s content.

\item \textbf{Weibo}~\cite{Weibo}: Weibo dataset is a user-user interaction network-based directed graph dataset on the common hashtag from Tencent-Weibo (A twitter-like platform in China). Suspicious users are marked as anomalies based on the temporal information of their posts. The node attribute of each user is based on the location information of the post and the bag-of-words representation of the post's content.

\item \textbf{Reddit}~\cite{reddit1,reddit2}: Reddit is a user-subreddit interaction network on the social media platform Reddit. The users which are banned from the subreddit are marked as anomalous users. The post of users and subreddits are converted into feature vectors representing the LIWC category~\cite{pennebaker2001linguistic} and the summation of these vectors are the attribute of the user and subreddit.

\item \textbf{Disney}~\cite{disney} and \textbf{Books}~\cite{booksandenron}: Disney and books are co-purchase networks of movies and books. The anomaly label of Disney is obtained by majority voting from the school students whereas the books dataset’s anomaly label is determined by the amazon fail tag information. Both datasets' feature contains information about price, number of reviews, and ratings.

\item \textbf{Enron}~\cite{booksandenron}: Enron is an email-interaction network where the nodes are email addresses and edges represent the interactions between email addresses. The email address that sends spam emails is marked as anomalies. The feature information of email addresses is given by the average length of the email body, the average number of recipients, and the time gap between two emails. 
\end{itemize}

\end{document}